\documentclass[paper=letter,DIV=10]{scrartcl}

\usepackage[utf8]{inputenc}
\usepackage{amsmath,amsfonts,amssymb,amsthm,bm,mathtools}
\usepackage{xcolor}
\usepackage{hyperref}
\usepackage{enumitem}
\usepackage[capitalise]{cleveref}
\usepackage{indentfirst}
\usepackage{booktabs}
\usepackage{siunitx}
\usepackage{multirow}
\usepackage{float}
\usepackage{graphicx}
\usepackage{subcaption}
\usepackage{tabularx}
\usepackage{makecell}
\usepackage{comment}
\usepackage{authblk}

\DeclareMathOperator*{\argmax}{arg\,max}
\DeclareMathOperator*{\argmin}{arg\,min}
\DeclarePairedDelimiterX{\infdivx}[2]{(}{)}{%
  #1\;\delimsize\|\;#2%
}
\newcommand{\DKL}{D_{\text{KL}}\infdivx}

\setkomafont{title}{\sffamily}
\setkomafont{author}{\normalsize}
\setkomafont{date}{\normalsize}

\title{Solving High-dimensional Inverse Problems Using Amortized Likelihood-free Inference with Noisy and Incomplete Data}
\author[a]{Jice Zeng}
\author[b]{Yuanzhe Wang}
\author[a,b]{Alexandre M. Tartakovsky}

\author[a]{David Barajas-Solano\thanks{Corresponding author: David.Barajas-Solano@pnnl.gov (David Barajas-Solano)}}

\affil[a]{Physical and Computational Sciences Directorate, Pacific Northwest National Laboratory, Richland, 99354, WA, United States of America}
\affil[b]{Department of Civil and Environmental Engineering, University of Illinois Urbana-Champaign, Urbana, IL 61801, United States of America}

\date{} 

\begin{document}
\maketitle
\begin{abstract}

We present a likelihood-free probabilistic inversion method based on normalizing flows for high-dimensional inverse problems. The proposed method is composed of two complementary networks: a summary network for data compression and an inference network for parameter estimation.
The summary network encodes raw observations into a fixed-size vector of summary features, while the inference network generates samples of the approximate posterior distribution of the model parameters based on these summary features.
The posterior samples are produced in a deep generative fashion by sampling from a latent Gaussian distribution and passing these samples through an invertible transformation.
We construct this invertible transformation by sequentially alternating conditional invertible neural network and conditional neural spline flow layers.
The summary and inference networks are trained simultaneously.

We apply the proposed method to an inversion problem in groundwater hydrology to estimate the posterior distribution of the log-conductivity field conditioned on spatially sparse time-series observations of the system's hydraulic head responses.
The conductivity field is represented with 706 degrees of freedom in the considered problem.
The comparison with the likelihood-based iterative ensemble smoother PEST-IES method demonstrates that the proposed method  accurately estimates the parameter posterior distribution and the observations' predictive posterior distribution at a fraction of the inference time of PEST-IES.
\end{abstract}

\section{Introduction}
\label{sec:Introduction}

Inverse problems are key for characterizing parameterized models of physical systems.
Unlike forward problems, which predict outcomes given model parameters, inverse problems work in the reverse direction, seeking to estimate model parameters or system states from partial measurements of observable quantities \cite{tarantola2005inverse}.
Inverse problems allow us to update computational models
to better align with observed data, thereby enhancing the models' predictive capabilities \cite{neto2012introduction}. Application areas leveraging inverse problems include medical imaging \cite{aguilo2010inverse}, geophysics \cite{bilionis2014crop}, structural health monitoring \cite{zeng2023bayesian}, reservoir engineering \cite{padmanabha2021solving}, and astronomy \cite{bellinger2020inverse}, among others.

Solving inverse problems introduces significant challenges. First, measurements are often sparse and contaminated by noise. Second, the inherent ill-posedness of inverse problems might lead to non-uniqueness and instability of the solutions. Finally, inverse solutions require multiple evaluations of the forward models, which might be unfeasible due to the high computational cost  \cite{karumuri2023learning}. There are two primary approaches to tackle inverse problems: (i) deterministic methods and (ii) Bayesian inference (BI) \cite{calvetti2018inverse}. Deterministic methods cast the parameter estimation problem as an optimization problem that seeks to minimize the discrepancy between model predictions and observed data. To mitigate ill-posedness, regularization techniques are frequently employed, resulting in well-posed optimization problems, and the quality of the parameter reconstructions depends strongly on the regularizer choice. Deterministic methods typically yield only point estimates of model parameters and do not account for uncertainties arising from modeling errors and measurement noise.

On the other hand, BI estimates a distribution of the model parameters by treating them as random variables. The starting point of BI is the prior distribution of model parameters, which acts as a form of regularization. Bayes' theorem is then employed to merge prior knowledge with observation data encoded into the data likelihood, resulting in a posterior distribution \cite{stuart2010inverse}. Unlike deterministic methods, BI by construction quantifies uncertainty in the parameter estimates. Therefore,
 BI is especially valuable for problems with significant measurement noise, non-uniqueness, and model uncertainty, providing a comprehensive perspective on potential models and their associated confidence levels \cite{dashti2013bayesian}.

BI methods can be categorized as follows based on their approach to evaluating the likelihood function that links model parameters to observations: (i) likelihood-based inference and (ii) likelihood-free inference---also known as simulation-based inference. Likelihood-based inference estimates the posterior distribution by explicitly evaluating the likelihood function, which should be explicitly defined and tractable. The gold standard for numerically sampling the posterior distribution is the Markov chain Monte Carlo (MCMC) method.
While conceptually straightforward, likelihood-based inference poses various challenges. First, directly evaluating the likelihood function is often either intractable or unfeasible \cite{lueckmann2021benchmarking} as in the case of multi-level models where sub-models are hierarchically connected \cite{zeng2023recursive}, continuous-time state-space models \cite{durbin2012time}, and nonlinear dynamic models \cite{huang2019state}, among others. Second, likelihood-based inference requires the repeated evaluation of the forward model to ensure convergence to the true posterior, which can be computationally intractable depending on the computational cost of the forward model. Third, likelihood-based inference methods are often hampered by the curse of dimensionality, which is potentially challenging for problems with spatially heterogeneous parameters, which often demand high-dimensional characterizations \cite{yeung2024conditional}. While the use of surrogate models, such as Gaussian processes \cite{binois2022survey,zeng2023machine}, polynomial chaos expansion \cite{marzouk2009dimensionality,yan2019adaptive}, and deep neural network \cite{mo2019deep,li2020nett} models, can reduce the computational cost, these methods still require likelihood evaluation and introduce additional epistemic uncertainty. Moreover, constructing accurate surrogate models becomes particularly problematic when dealing with high-dimensional inputs and outputs.

In contrast, likelihood-free inference aims to estimate the posterior distribution without directly evaluating the likelihood function
and relying solely on synthetic data generated for various parameter choices.
Approximate Bayesian computation (ABC) is perhaps the most developed form of likelihood-free inference \cite{sunnaaker2013approximate}. In ABC, parameters are repeatedly sampled from a prior distribution, and synthetic data is generated by running the forward model using these sampled parameters. Parameters whose simulated data sufficiently resemble actual observations, as decided by user-defined criteria, are retained as samples of the target posterior; others are discarded. However, ABC has several disadvantages resulting from its accept-reject mechanism: it is challenging to set reasonable criteria for accurate posterior estimation. Additionally, the entire ABC process must be restarted with new measurement datasets, limiting its application to inference for a single set of observed data \cite{radev2020bayesflow}.

In recent developments, deep generative models (DGMs) have emerged as a promising approach for solving inverse problems via likelihood-free inference. Central to DGMs is the strategy of a nonlinear map from a latent distribution, such as a Gaussian, to a complex target distribution. Consequently, approximating the posterior can be done by learning this nonlinear transformation. Such a learning process can be efficiently executed using neural networks trained via back propagation \cite{liu2021density}. After sufficient training, one can draw samples from a straightforward latent distribution and produce posterior distribution samples by applying the learned transformation. Numerous studies have explored the use of DGMs for inverse problems,
including applications of generative adversarial networks (GANs) \cite{grover2018flow,patel2022solution} and variational autoencoders \cite{laloy2017inversion,canchumuni2019towards}.
GANs and their variants, in particular, have become increasingly popular for likelihood-free inference.
Baptista et al. \cite{baptista2020conditional} introduced conditional sampling for likelihood-free inference using a monotone GAN, which learns a block triangle transport map between a reference and target measure through adversarial learning combined with a monotonicity penalty. Ramesh et al. \cite{ramesh2022gatsbi} developed a conditional GAN to redefine the variational objective for learning an implicit posterior distribution over latent variables, demonstrating significant promise for high-dimensional inverse problems. Additionally, Patel et al. \cite{patel2019bayesian} proposed using a GAN to model a field characterized by numerous discrete representations, where the GAN’s generator maps a latent distribution to the target distribution for the field. Despite the significant impact of GANs on posterior estimation, they are notoriously challenging to train, highly sensitive to hyperparameter settings, and prone to the issue of mode collapse \cite{borji2022pros}.

An alternative approach to DGM-based likelihood-free inference is the use of flow-based models, particularly normalizing flows (NFs) \cite{papamakarios2021normalizing}.
NFs use invertible networks, allowing direct sampling from complex, parameterized probability distributions and precise density evaluation using a change-of-variables formula. 
NFs have demonstrated strong performance in high-quality posterior estimation across various fields such as economics \cite{shiono2021estimation}, epidemiology \cite{radev2021outbreakflow}, physics \cite{kang2023noise}, computer vision \cite{ardizzone2019guided}, and civil engineering \cite{zeng2023probabilistic}. Additionally, they provide precise log-likelihood evaluations and avoid the training stability issues often encountered in GANs.

In this study, we introduce a likelihood-free inference approach using NFs for estimating high-dimensional spatially heterogeneous parameters in initial-boundary value problems (IBVPs). The proposed model consists of two complementary deep neural networks: a summary network and an inference network. The summary network compresses raw time-series observation data employing one-dimensional convolutions in time, extracting the most informative features, moving beyond manually crafted summary statistics. The inference network is constructed by alternating between two types of flow-based invertible layers: affine coupling layers (ACLs) and spline layers (SLs). By alternating between these two types of layers we enhance the model’s expressivity and flexibility.
The summary network forwards the summarized observations to the inference network, thereby conditioning the entire NF on these observations. The inference and summary networks are jointly trained through backpropagation using synthetic data. This concurrent training ensures that the data representation is adequate for estimating the posterior distributions of the parameters.
We apply our approach to estimating a hydraulic log-conductivity field in a numerical groundwater model with 706 special computational cells using noisy time-dependent observations of the hydraulic head at 13 sparse locations.
The estimation results are compared with the likelihood-based iterative ensemble smoother (IES) method implemented in the PEST software in terms of both accuracy and computational efficiency. The main advantages of the proposed method are as follows:
\begin{enumerate}
    \item It can accurately estimate high-dimensional parameters without the need to evaluate the likelihood function and make any assumptions about the posterior.
    \item Unlike traditional methods such as MCMC or ABC, it supports amortized inference, i.e., it allows incorporating new measurements without retraining the model, which enables near real-time predictions with streaming data.
\end{enumerate}

\section{Background}
\subsection{Inverse problem formulation} \label{sec:inverse formulation}

The motivating application is estimation of spatially heterogeneous parameters in groundwater flow models. In this work, we consider two-dimensional groundwater flow over the simulation domain $\Omega \in \mathbb{R}^2$ governed by the nonlinear partial differential equations
\begin{gather}\label{eq:pde1}
S_y \frac{\partial u (\mathbf{x},t) }{\partial t} = \nabla \cdot [\kappa(\mathbf{x}) u(\mathbf{x},t) \nabla u(\mathbf{x},t)] - r(\mathbf{x},t), \quad \mathbf{x} \in \Omega, \, t > 0 
\end{gather}
subject to appropriate initial and boundary conditions. 
Here, \( u(\mathbf{x}, t) \) denotes the hydraulic head field, $\kappa(\mathbf{x})$ the spatially varying hydraulic conductivity, $r(\mathbf{x},t)$ the source term, and $S_y$ the (assumed to be known) specific yield.

For a given $\kappa(\mathbf{x})$, Eq \eqref{eq:pde1} is solved using the control volume finite-difference solver \textsc{MODFLOW} \cite{hughes2022documentation}. 
In the numerical model, the domain is discretized with $N$ cells and the conductivity field is represented with the vector $\mathbf{\kappa} \in \mathbb{R}^{N}$ of its values at the cell centers. 
We solve the inverse problem for $\mathbf{y} = \ln \bm{\kappa} \in \mathbb{R}^{N}$ rather than  $\mathbf{\kappa}$, which guarantees that  $\kappa(\mathbf{x})$ is strictly positive. 

We assume that only hydraulic head responses \( \mathbf{u} \), collected at $N_u$ sensor locations ($N_u<N$), are available for solving the inverse problem.
Because the number of observations is smaller than the number of parameters, this inverse problem is, in general,  ill-posed.
We note that many inverse methods reduce the dimensionality of the parameter space because of the high computational cost of finding the inverse solution in the full parameter space, an approximation leading to some errors.  
Here, we formulate the inverse problem in the original  $\mathbb{R}^{N}$ space, which makes it possible to solve problems where the dimension reduction is not possible, i.e., when all parameters are uncorrelated and equally important. 

The proposed likelihood-free inference model is trained using a synthetic labeled dataset consisting of pairs of hydraulic head values at the sensor locations and $\mathbf{y}$. The $\mathbf{y}$ vectors are generated by sampling the log-conductivity prior distribution, and the corresponding hydraulic head field is computed with \textsc{MODFLOW}.

\subsection{Bayesian inference} \label{sec:Bayesian inference}

Let $\bm{\theta} \coloneqq \mathbf{y}$ denote the uncertain model parameters.
Furthermore, let $\mathcal{M}_k(\bm{\theta})$ denote the time-series forward model's output at timestep $k$ consisting of the hydraulic head at $N_u$ observation locations.
We also consider the excitations, e.g., boundary conditions, initial conditions, and body forces, as fixed in this setting.
We model the time-series observations of hydraulic head at sensor locations via the additive noise model 
\begin{equation}\label{headmodel}
\mathbf{u}_{o,k} \coloneqq \mathcal{M}_k(\bm{\theta}) + \bm{\epsilon}_k.
\end{equation}
The term \( \bm{\epsilon}_k \sim \mathcal{N}(0, \bm{\Sigma})\) denotes zero-mean random Gaussian noise with covariance matrix \(\bm{\Sigma} \coloneqq \operatorname{diag} [ \sigma_1^2, \sigma_2^2, \ldots, \sigma_{N_u}^2] \in \mathbb{R}^{\mathrm{N}_u \times \mathrm{N}_u}\), in which \(\sigma_i^2\), \(i = 1, 2, \ldots, N_u\) is the variance of measurements at the $i$th observation location.
Furthermore, we assume that measurement errors are uncorrelated in time, that is, $\mathbb{E} [ \bm{\epsilon}_{p} \bm{\epsilon}^{\top}_q] = \partial_{pq} \boldsymbol{\Sigma}$, where $\delta_{pq}$ denotes the Kronecker delta.

The posterior distribution of $\bm{\theta}$, i.e., the distribution of $\bm{\theta}$ conditional on the observations $\mathbf{u}_{o, 1:k}$, is given by Bayes' theorem as
\begin{equation}\label{Bayes}
p(\bm{\theta} | \mathbf{u}_{o,1:k}) = \frac{p(\mathbf{u}_{o,1:k} | \bm{\theta}) \, p(\bm{\theta})}{\int p(\mathbf{u}_{o,1:k} | \bm{\theta}) \, p(\bm{\theta}) \, d\bm{\theta}} \propto p(\bm{\theta}) \, p(\mathbf{u}_{o,1:k} | \bm{\theta})
\end{equation}
where \( p_{\theta}(\bm{\theta}) \) denotes the prior distribution of the model parameters, which encapsulates our initial beliefs before data acquisition. The matrix \( \mathbf{u}_{o,1:k} = [\mathbf{u}_{o,1}, \mathbf{u}_{o,2}, \ldots, \mathbf{u}_{o,k}] \), with each \( \mathbf{u}_{o,i} \in \mathbb{R}^{N_u} \) for \( i = 1, 2, \ldots, k \), represents the sequence of collected measurements from time \( t_1 \) to \( t_k \). The likelihood function \( p_{\mathbf{u}_{o,1:k} | \theta}(\mathbf{u}_{o,1:k} | \bm{\theta}) \) quantifies the plausibility of the observed sequence given the model parameters \(\bm{\theta}\). This likelihood function is directly tied to the error model, which characterizes the discrepancies between the predicted and observed data. The error model enables us to account for measurement uncertainties or model imperfections, ensuring that the likelihood appropriately reflects the real-world observations.
This function plays a crucial role in updating our initial beliefs in light of new evidence from the data. The denominator, known as the marginal likelihood or evidence, integrates over all possible values of \(\bm{\theta}\), thus normalizing the product of the prior distribution and the likelihood, ensuring the posterior distribution is a valid probability distribution. The Bayesian framework facilitates a systematic update of our knowledge about \(\bm{\theta}\) as new data \( \mathbf{u}_{o,1:k} \) is observed, effectively integrating the prior information with empirical evidence.

The posterior distribution \( p(\bm{\theta} | \mathbf{u}_{o,1:k}) \) can be estimated through various Bayesian inference techniques, including MCMC sampling methods and ABC. However, these methods often face challenges such as high computational cost and inefficiency in high-dimensional parameter spaces, as discussed in Sec.~\ref{sec:Introduction}. We propose a likelihood-free inference approach to address these limitations.
Unlike most existing likelihood-free inference approaches, which are typically limited to low-dimensional parameter spaces (e.g., a few or up to a dozen parameter dimensions), our approach leverages the capacity of NFs to infer high-dimensional parameter fields, specifically a 706-dimensional log-conductivity field, using sparse and noisy time-series observations. Our model employs two types of invertible layers---ACLs and SLs---to enhance estimation accuracy. The proposed framework effectively tackles the challenges of estimating high-dimensional, spatially distributed parameter fields in an amortized fashion as described in the following section.

\section{Methodology}\label{sec:methodology}

\begin{figure}[ht]
    \centering
    \includegraphics[width=\textwidth]{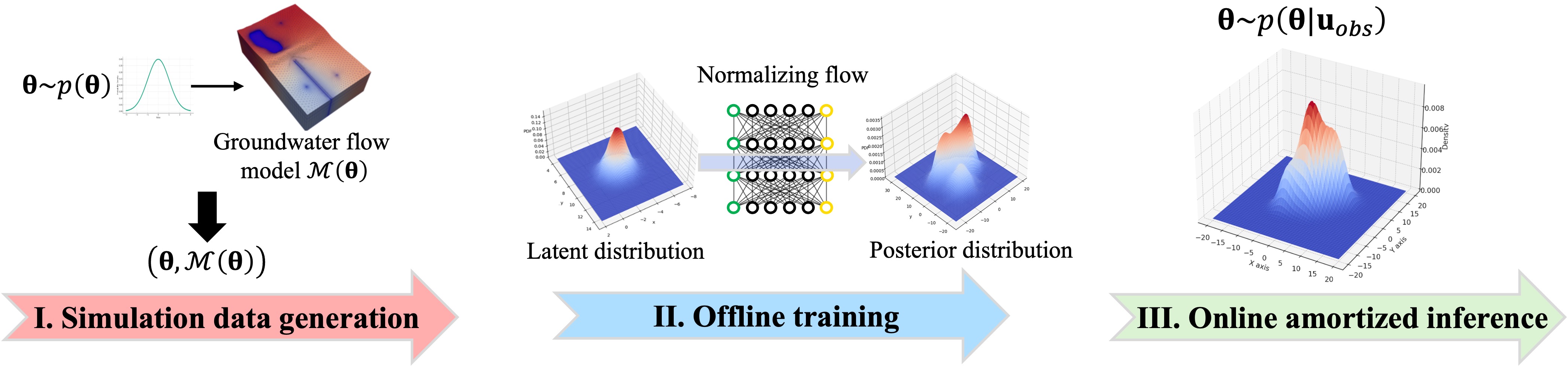} 
    \caption{The workflow of the proposed method for Bayesian model inversion.}
    \label{fig:1}
\end{figure}

The proposed likelihood-free method for parameter estimation consists of three distinct phases, as depicted in Fig. \ref{fig:1}.
In Phase I, simulation data are generated by sampling model parameters from the prior distribution, \( \bm{\theta} \sim p(\bm{\theta}) \), and using them to run a forward model. This process produces a training dataset of \( M \) pairs, \( \{ (\bm{\theta}^{(i)}, \mathbf{u}^{(i)}_{\text{syn},1:k}) \}_{i=1}^M \), where \( \bm{\theta}^{(i)} \) represents the sampled parameters and \( \mathbf{u}^{(i)}_{\text{syn},1:k} \) denotes the corresponding simulated observations over the time steps \( 1:k \).
Phase II consists of offline training the NF model using simulation data.
Finally, Phase III consists of online amortized inference, in which we generate samples from the posterior distribution for any observation dataset.
The term ``online amortized inference'' refers to the fact that in our methodology, the intensive computational effort of data simulation and model training are front-loaded during Phases I and II.
Consequently, inference can be performed by sampling from the latent simple distribution and the learnt transformation without starting anew for different sets of observation data.
This confers a substantial boost in efficiency for applications in which it is necessary to estimate varying parameters (such as boundary conditions) from new observations.

It is worth mentioning that the need for the retraining flow model depends on the nature of the changes to $N_u$ (number of observation locations) and $k$ (number of time steps). If the observations contain fewer time steps than the training data, the model does not need to be retrained and can still be used to infer parameters, given that we employ one-dimensional convolutions to summarize the time-series data. If the number of time steps in the observations exceeds those in the training data, the model can also be reused by employing recursive likelihood-free inference \cite{zeng2023recursive}. On the other hand, if the number of observation locations differs from that in the training data, retraining the model is necessary. The spatial arrangement of observations directly affects the structure of the likelihood-free inference process and the mappings learnt by the NF model.

\subsection{Likelihood-free inference by normalizing flow}\label{sec:NF}

The proposed likelihood-free inference framework is based on NFs and integrates concepts from Bayesian inference and NF theory. Originally developed by Tabak and Turner \cite{tabak2013family}, and further by Tabak and Vanden-Eijnden \cite{tabak2010density}, NFs were primarily applied to feature extraction and density estimation. NFs are generative models that constructs complex distributions by transforming a known simple probability distribution (e.g., a multivariate normal distribution) through a series of invertible and differentiable transformations.

\begin{figure}[ht]
    \centering
    \includegraphics[width=0.95\textwidth]{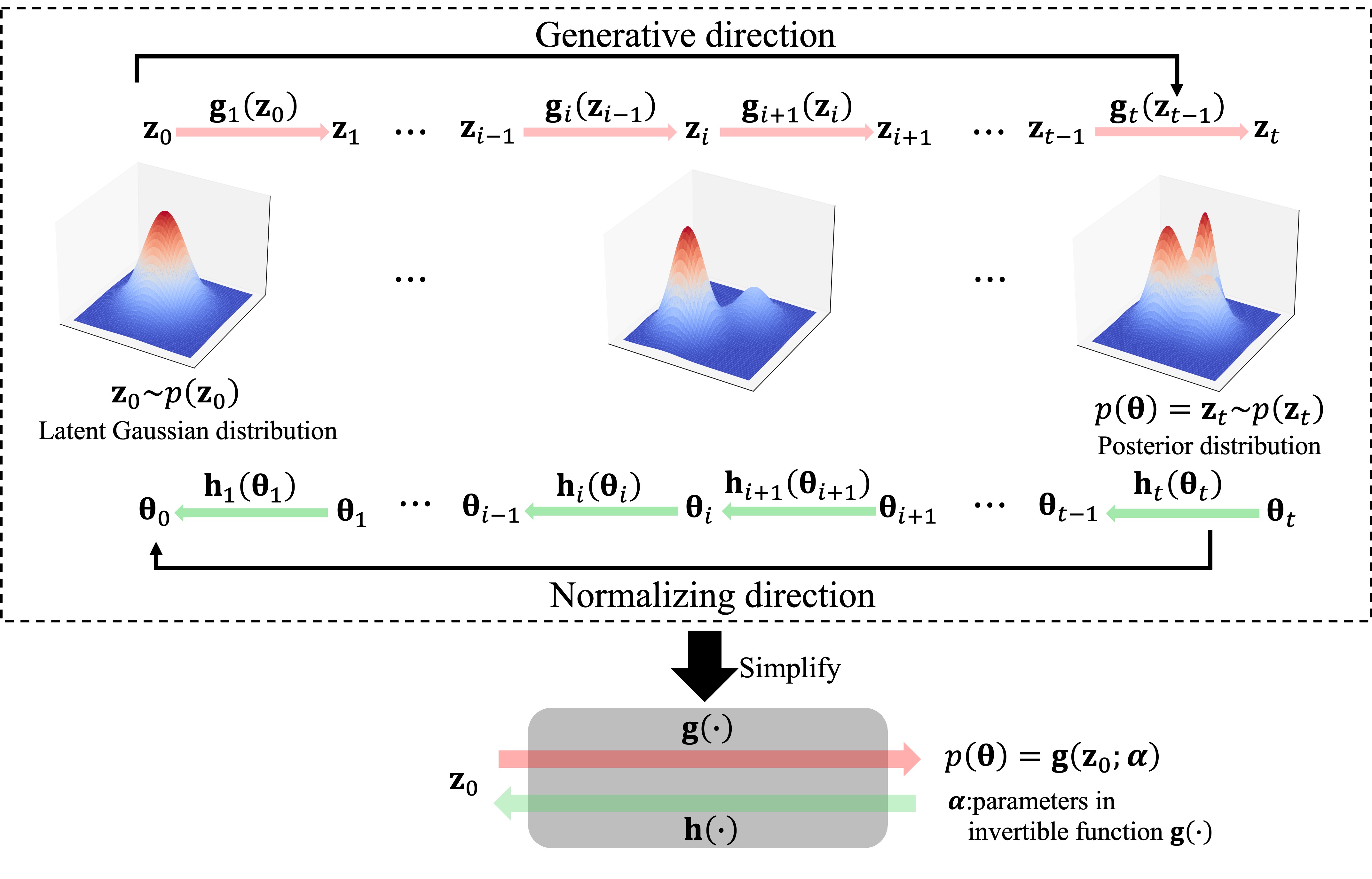} 
    \caption{Schematic of normalizing flow.}
    \label{fig:2}
\end{figure}

Fig. \ref{fig:2} demonstrates the principles of NF. There are two key forms of mapping between the base (latent) distribution and the target distribution: the \textbf{generative direction} and the \textbf{normalizing direction}. In the generative direction, reversible transformations convert a simple base distribution into a complex target distribution. Conversely, in the normalizing direction, the complex and irregular target distribution \( p(\bm{\theta}) \) is transformed back into the structured base distribution \( p(\mathbf{z}_0) \), which is typically chosen as a standard Gaussian distribution \( \mathcal{N}(\mathbf{z}_0 | \bm{0}, \mathbf{I}) \). 

NF consists of a sequence of \( t \) transformations applied to latent variables \( \mathbf{z} \) as follows:
\begin{equation}\label{zchain}
\mathbf{z} \coloneqq \mathbf{z}_0 \to \mathbf{z}_1 \to \mathbf{z}_2 \to \dots \to \mathbf{z}_t \coloneqq \bm{\theta}
\end{equation}
where each transformation \( \mathbf{z}_i = \mathbf{g}_i(\mathbf{z}_{i-1}) \) is a bijective mapping. Importantly, the transformations \( \mathbf{g}_i \) must satisfy the following properties:
\begin{enumerate}
    \item \textbf{Invertibility}: Ensures a one-to-one mapping between the base distribution and the target distribution, allowing for both directions (generative and normalizing) to be well-defined.
    \item \textbf{Computationally efficient Jacobian}: The Jacobian determinant \( \left| \det \left( \frac{\partial \mathbf{g}_i}{\partial \mathbf{z}_{i-1}} \right) \right| \) should be easy to compute for tractable density estimation.
    \item \textbf{Dimensional consistency}: The dimensions of \( \mathbf{z}_i \) and \( \mathbf{z}_{i-1} \) must match to maintain a valid transformation.
\end{enumerate}

At each step, the distribution \( p(\mathbf{z}_t) \) is related to \( p(\mathbf{z}_{t-1}) \) through the change of variables formula
\begin{equation} \label{zchange-rule}
p(\mathbf{z}_t) = p(\mathbf{z}_{t-1}) \left| \det \left( \frac{\partial \mathbf{z}_t}{\partial \mathbf{z}_{t-1}} \right)^{-1} \right| = \left| \det \left( \frac{\partial \mathbf{g}_t(\mathbf{z}_{t-1})}{\partial \mathbf{z}_{t-1}} \right) \right|^{-1}
\end{equation}
where \(\det(\cdot)\) is the determinant operator. Eq. \ref{zchange-rule} describes the relationship between the densities of \( \mathbf{z}_{t-1} \) and \(\mathbf{z}_t \) by employing the determinant of the Jacobian matrix of the transformation \( \mathbf{g}_t \). Specifically, the Jacobian matrix serves as a quantitative metric of how the transformation function \( \mathbf{g}_t \) compresses elements of the base distribution to align with the target distribution.

In the normalizing direction, we simplify the complex target distribution \( p(\bm{\theta}) \) by transforming it back to a base distribution. The full transformation from the target distribution \( p(\bm{\theta}) \) to the base distribution \( p(\mathbf{z}_0) \) is captured by the composite function \( \mathbf{h}(\cdot) \)
\begin{equation}\label{norm_direc}
\mathbf{z} = \mathbf{h}(\bm{\theta}) = \mathbf{h}_t \circ \mathbf{h}_{t-1} \circ \dots \circ \mathbf{h}_1(\bm{\theta})
\end{equation}
where \( \mathbf{h}_i \) are the inverse transformations of \( \mathbf{g}_i \). In the generative direction, we apply the sequence of transformations \( \mathbf{g}_i \) in reverse order to map from \( \mathbf{z}_0 \) to \( \bm{\theta} \)
\begin{equation}\label{gen_direc}
\bm{\theta} = \mathbf{g}_1 \circ \mathbf{g}_2 \circ \dots \circ \mathbf{g}_t(\mathbf{z}_0).
\end{equation}
Thus, the distribution \( p(\bm{\theta}) \) can be expressed in terms of the base distribution \( p(\mathbf{z}_0) \) and the Jacobian determinant of the composite transformation \( \mathbf{h}(\bm{\theta}) \). The complex target distribution is thus approximated by the transformed latent distribution \( p(\mathbf{z}_0) \) using the change-of-variables rule. The distribution \( p(\bm{\theta}) \) is given by
\begin{equation}\label{unconditional}
p(\bm{\theta}) = p(\mathbf{z}_0) \left| \det \left( \frac{\partial \mathbf{h}(\bm{\theta})}{\partial \bm{\theta}} \right) \right| = p(\mathbf{z}_0) \left| \det \left( \frac{\partial \mathbf{g}(\mathbf{h}(\bm{\theta}))}{\partial \mathbf{z}_0} \right) \right|^{-1}.
\end{equation}

To compute the Jacobian of the composite transformation \( \mathbf{h}(\bm{\theta}) \), we apply the chain rule. The Jacobian of the entire transformation is obtained by multiplying the Jacobians of each individual transformation step
\begin{equation}\label{jacobian}
\frac{\partial \mathbf{g}(\mathbf{h}(\bm{\theta}))}{\partial \mathbf{z}} = \frac{\partial \mathbf{g}_t}{\partial \mathbf{z}_{t - 1}} \frac{\partial \mathbf{g}_{t - 1}}{\partial \mathbf{z}_{t - 2}} \cdots \frac{\partial \mathbf{g}_1}{\partial \mathbf{z}_0}
\end{equation}
which shows how the Jacobian matrix at each step contributes to the overall Jacobian of the transformation. The composition of bijective functions ensures that the final transformation is also bijective, and the chain rule allows us to efficiently compute the Jacobian determinant across multiple layers of transformations.

Eq. \ref{unconditional} can be easily generalized to a conditional NF framework, aiming to learn the conditional posterior distribution \( p(\bm{\theta} | \mathbf{u}_{o,1:k}) \), where \( \mathbf{u}_{o,1:k} \) is the time-series data collected at \( k \) time steps. The aim of conditional NF is to approximate the target posterior distribution \( p(\bm{\theta} | \mathbf{u}_{o,1:k}) \) of \( \bm{\theta} \) given any observations \( \mathbf{u}_{o,1:k} \), utilizing the parameterized posterior approximation
\begin{gather} \label{eq:8}
p(\bm{\theta}|\mathbf{u}_{o,1:k}) \approx p_{\bm{\alpha}}(\bm{\theta}|\mathbf{u}_{o,1:k}) \\
\label{eq:9}
p(\bm{\theta} | \mathbf{u}_{o,1:k}) = p(\mathbf{z}_0 | \mathbf{u}_{o,1:k}) \left| \det \left( \frac{\partial \mathbf{h}(\bm{\theta}; \mathbf{u}_{o,1:k})}{\partial \bm{\theta}} \right) \right| = p(\mathbf{z}_0 | \mathbf{u}_{o,1:k}) \left| \det \left( \frac{\partial \mathbf{g}(\mathbf{h}(\bm{\theta}; \mathbf{u}_{o,1:k}))}{\partial \mathbf{z}_0} \right) \right|^{-1}
\end{gather}
where \( p_{\bm{\alpha}}(\bm{\theta}|\mathbf{u}_{o,1:k}) \) is given by \eqref{eq:8} for a normalizing transformation $\mathbf{h}_{\bm{\alpha}}$.
Samples from this approximate posterior are drawn as
\begin{equation} \label{eq:10}
\bm{\theta} \sim p_{\bm{\alpha}}(\bm{\theta}|\mathbf{u}_{o,1:k}) \Leftrightarrow \bm{\theta} = \mathbf{h}_{\bm{\alpha}}^{-1}(\mathbf{z};\mathbf{u}_{o,1:k}), \text{ with } \mathbf{z} \sim \mathcal{N}(\mathbf{z}|\bm{0}, \mathbf{I}).
\end{equation}

In practice, the design of invertible functions \( \mathbf{g}(\cdot) \) for nonlinear bijective mappings poses significant challenges. Commonly, these functions are parameterized through layers of invertible neural networks, each providing efficient bijections with cost-effective Jacobian evaluations. A detailed review of such architectures is available in \cite{kobyzev2020normalizing}. In this work we employ ACLs and SLs as the fundamental elements of the invertible function. These concepts are further explained in Section 3.3. 

The conditional NF is trained by finding the optimal parameter \( \hat{\bm{\alpha}} \)
that minimizes the Kullback-Leibler (KL) divergence between the target and approximate posterior across all observation data \( \mathbf{u}_{o,1:k} \). The optimal \( \hat{\bm{\alpha}} \) is given by \cite{radev2020bayesflow}
\begin{equation}
  \label{eq:KLD_NF}
  \begin{split}
    \hat{\bm{\alpha}} &= \argmin_{\bm{\alpha}} \, \mathbb{E}_{p(\mathbf{u}_{o,1:k})} \left[ \DKL{p(\bm{\theta}|\mathbf{u}_{o,1:k})}{p_{\bm{\alpha}} (\bm{\theta}|\mathbf{u}_{o,1:k})} \right], \\
                      &= \argmin_{\bm{\alpha}} \, \mathbb{E}_{p(\mathbf{u}_{o,1:k})} \left[ \mathbb{E}_{p(\bm{\theta}|\mathbf{u}_{o,1:k})} \left[ \log p(\bm{\theta}|\mathbf{u}_{o,1:k}) - \log p_{\bm{\alpha}} (\bm{\theta}|\mathbf{u}_{o,1:k}) \right] \right], \\
                      &= \argmax_{\bm{\alpha}} \, \mathbb{E}_{p(\mathbf{u}_{o,1:k})} \left[ \mathbb{E}_{p(\bm{\theta}|\mathbf{u}_{o,1:k})} \left[ \log p_{\bm{\alpha}} (\bm{\theta}|\mathbf{u}_{o,1:k}) \right] \right], \\ 
                      &= \argmax_{\bm{\alpha}} \, \iint p(\mathbf{u}_{o,1:k}, \bm{\theta}) \log p_{\bm{\alpha}} (\bm{\theta}|\mathbf{u}_{o,1:k}) \, d\mathbf{u}_{o,1:k} \, d\bm{\theta}.
  \end{split}
\end{equation}
where \( \mathbb{E}(\cdot) \) denotes the expectation operator, and \( \DKL{\cdot}{\cdot} \) represents the KL divergence.
Here we ignore the term \( \log p(\bm{\theta} | \mathbf{u}_{o,1:k}) \) because it is independent of \( \bm{\alpha }\).
Following Eq. \ref{eq:9}, we write \( p_{\bm{\alpha}} (\bm{\theta}|\mathbf{u}_{o,1:k}) = p(\mathbf{z} = \mathbf{h}_{\bm{\alpha}} (\bm{\theta};\mathbf{u}_{o,1:k})) \left | \det \partial \mathbf{h}_{\bm{\alpha}} (\bm{\theta};\mathbf{u}_{o,1:k}) / \partial \bm{\theta} \right| \). Consequently, the above optimization problem can be rewritten as
\begin{equation}\label{eq:objective}
\hat{\bm{\alpha}} = \argmax_{\bm{\alpha}} \iint p(\mathbf{u}_{o,1:k}, \bm{\theta}) \left [ \log p(\mathbf{h}_{\bm{\alpha}} (\bm{\theta}; \mathbf{u}_{o,1:k})) + \log |\det \bm{J}_{\mathbf{h}_{\bm{\alpha}}} | \right ] \, d\mathbf{u}_{o,1:k} \, d\bm{\theta}
\end{equation}
where \( \bm{J}_{\mathbf{h}_{\bm{\alpha}}} \) denotes the Jacobian matrix \( \partial \mathbf{h}_{\bm{\alpha}} (\bm{\theta}; \mathbf{u}_{o,1:k}) / \partial \bm{\theta} \). Given \( M \) pairs of simulated data \( \{(\bm{\theta}^{(i)}, \mathbf{u}_{1:k}^{(i)} )\}_{i=1}^M \), the expectation in Eq. \ref{eq:objective} can be approximated using Monte Carlo sampling as
\begin{equation} \label{eq:objective_empirical}
\hat{\bm{\bm{\alpha}}} = \argmax_{\bm{\bm{\alpha}}} \frac{1}{M} \sum_{i=1}^M \left [ \log p(\mathbf{h}_{\bm{\alpha}} (\bm{\theta}^{(i)}; \mathbf{u}_{1:k}^{(i)})) + \log |\det \bm{J}_{\mathbf{h}_{\bm{\alpha}}}^{(i)} | \right ].
\end{equation}

Recall that the latent variable \( \mathbf{z} \) follows an \(N\)-dimensional standard Gaussian distribution, so that
\begin{equation}
p(\mathbf{z} = \mathbf{h}_{\bm{\alpha}}(\bm{\theta}; \mathbf{u}_{o,1:k})) = \frac{1}{(2\pi)^{N/2}} \exp\left(-\frac{1}{2} \|\mathbf{h}_{\bm{\alpha}}(\bm{\theta}; \mathbf{u}_{o,1:k})\|^2 \right)
\end{equation}
Taking the negative logarithm leads to
\begin{equation}
-\log p(\mathbf{z} = \mathbf{h}_{\bm{\alpha}}(\bm{\theta}; \mathbf{u}_{o,1:k})) = \frac{N}{2} \log(2\pi) + \frac{1}{2} \|\mathbf{h}_{\bm{\alpha}}(\bm{\theta}; \mathbf{u}_{o,1:k})\|^2.
\end{equation}
Therefore, we can rewrite \eqref{eq:objective_empirical} as
\begin{equation}\label{eq:objective_final}
\hat{\bm{\alpha}} = \underset{\bm{\alpha}}{\text{argmin}} \, \mathcal{L}(\bm{\alpha}),
\end{equation}
where the loss function \( \mathcal{L}(\bm{\alpha}) \) is defined as
\begin{equation} \label{eq:objective_loss}
\mathcal{L}(\bm{\alpha}) = \frac{1}{M} \sum_{i=1}^M \left \{ \frac{1}{2} \left[\mathbf{h}_{\bm{\alpha}}(\bm{\theta}^{(i)}; \mathbf{u}^{(i)}_{o,1:k})\right]^2 - \log \left| \det \left( \frac{\partial \mathbf{h}_{\bm{\alpha}}(\bm{\theta}^{(i)}; \mathbf{u}^{(i)}_{o,1:k})}{\partial \bm{\theta}} \right) \right| \right \}.
\end{equation}
The minimization problem \eqref{eq:objective_final} can be solved using stochastic gradient descent algorithms. The second term in Eq. \ref{eq:objective_loss} reflects the log-volume change induced by the nonlinear transformation from the input space \( \bm{\theta} \) to the latent variable \( \mathbf{z} \), as captured by the determinant of the Jacobian. This adjustment ensures that the transformation's effect on the density is incorporated into the overall likelihood.

\subsection{Summary network} \label{sec:summary}

The summary network serves as a critical preprocessing step, which converts the $u$ measurements into a fixed, low-dimensional input (summary features) to the conditional NF network. This step is important because likelihood-free methods rely on the summary features capturing sufficient information about the measurements to ensure the inference quality.
While summary statistics have traditionally been manually crafted based on domain knowledge,
such choices may be unreliable, potentially losing important information and adversely affecting the inference process.

The adoption of a summary network offers several advantages, particularly in reducing the data dimensionality and improving the efficiency of NF training. First, it alleviates the computational burden by compressing high-dimensional raw data into a more manageable form, which enhances both training speed and precision. Secondly, preprocessing through the summary network helps remove redundancies and noise inherent in the raw data, which ensures that only the most relevant information is passed on to the inference stage. Finally, the summary network automates the extraction of informative features from the data, eliminating the need for manual feature selection \cite{radev2020bayesflow}.

The summary network consists of the parameterized transformation 
\begin{equation} \label{eq:21}
\hat{\mathbf{u}} = \mathcal{\bm{F}}_{\bm{\beta}}(\mathbf{u}_{o,1:k}) \
\end{equation}
where \( \hat{\mathbf{u}} \) represents the learned summary features from the raw data \( \mathbf{u}_{o,1:k} \), which will be utilized as input to the conditional NF.
The architecture of the summary network is chosen based on the characteristics of the observation data. For instance, bidirectional long short-term memory networks are especially suited for time-series data, enabling the effective handling of sequential measurements that exhibit long memory and nonlinear characteristics. Here we use a one-dimensional convolutional neural network to learn summary features.

The hyperparameters of conditional NF and the summary network are jointly optimized by employing the approximation $p(\bm{\theta} | \mathbf{u}_{o, 1:k}) \approx p_{\bm{\alpha}}(\bm{\theta} | \hat{\mathbf{u}})$ and minimizing the KL divergence between this approximation and the true posterior. Employing this approximation instead of \eqref{eq:9}, \eqref{eq:KLD_NF} becomes
\begin{equation} \label{eq:objective_joint}
(\hat{\bm{\alpha}}, \hat{\bm{\beta}}) = \argmax_{\bm{\alpha}, \bm{\beta}} \, \mathbb{E}_{p(\mathbf{u}_{o,1:k})} \left[ \mathbb{E}_{p(\bm{\theta}|\mathbf{u}_{o,1:k})} \left[ \log p_{\bm{\alpha}} (\bm{\theta}|\mathcal{\bm{F}}_{\bm{\beta}} (\mathbf{u}_{o,1:k})) \right] \right],
\end{equation}
and the Monte Carlo approximation \eqref{eq:objective_final} becomes
\begin{equation} \label{eq:objective_joint_final}
(\hat{\bm{\alpha}}, \hat{\bm{\beta}}) = \argmin_{\bm{\alpha}, \bm{\beta}} \, \mathcal{L}(\bm{\alpha}, \bm{\beta})
\end{equation}
with loss function
\begin{equation} \label{eq:objective_joint_loss}
\mathcal{L}(\bm{\alpha}, \bm{\beta}) = \frac{1}{M} \sum_{i=1}^M \left \{ \frac{1}{2} \left[ \mathbf{h}_{\bm{\alpha}} \left( \bm{\theta}^{(i)};\mathcal{\bm{F}}_{\bm{\beta}} \left( \mathbf{u}_{o,1:k}^{(i)} \right) \right) \right]^2 - \log \left| \det \left( \frac{\partial \mathbf{h}_{\bm{\alpha}} \left( \bm{\theta}^{(i)}; \mathcal{\bm{F}}_{\bm{\beta}} \left( \mathbf{u}^{(i)}_{o,1:k} \right) \right)}{\partial \bm{\theta}} \right) \right| \right \}.
\end{equation}

\subsection{Inference network} \label{sec:inference}

The core element of NFs is invertible transformations. In our study, we employ two families of parameterized invertible transformations: conditional ACLs (cACLs), and conditional SLs (cSLs). In this section, we present an inference network combining cACLs and cSLs that is capable of bidirectional mapping between posterior and latent distributions. This dual mapping capability is critical for enabling the forward transformation as well as the reverse propagation.

\subsubsection{Conditional affine coupling layers}

\begin{figure}[ht]
    \centering
    \begin{subfigure}[t]{\textwidth}
        \centering
        \includegraphics[width=\textwidth]{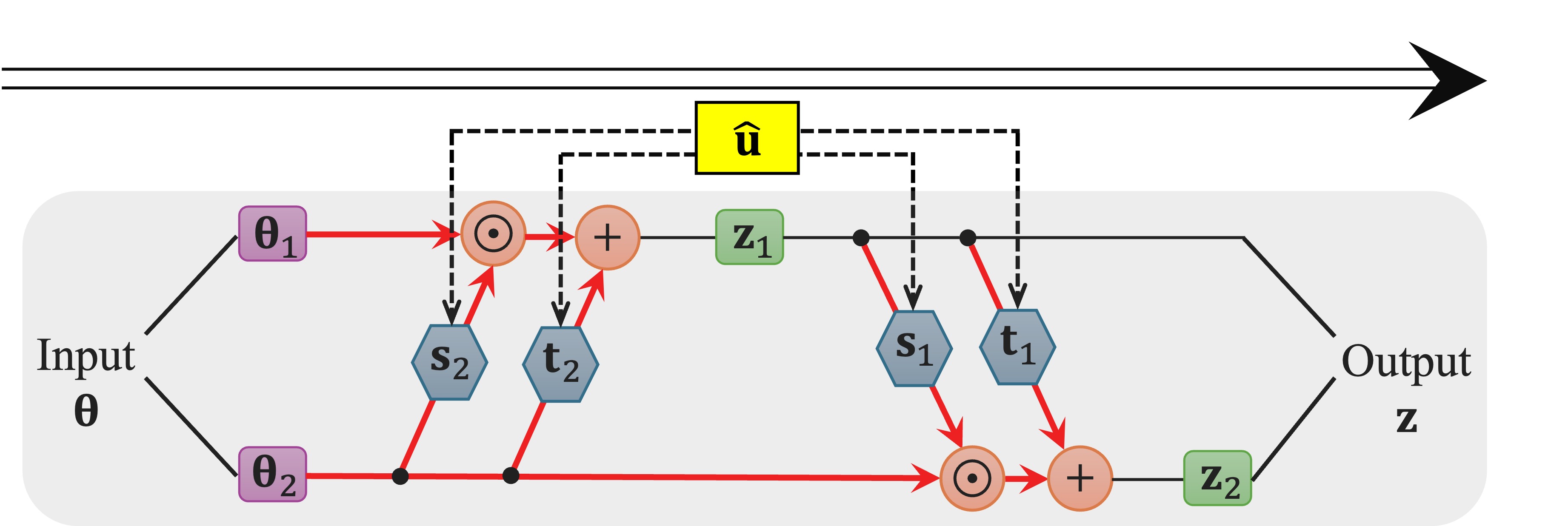}
        \caption{ forward transformation}
        \label{fig:3a}
    \end{subfigure}
    \vskip\baselineskip 
    \begin{subfigure}[t]{\textwidth}
        \centering
        \includegraphics[width=\textwidth]{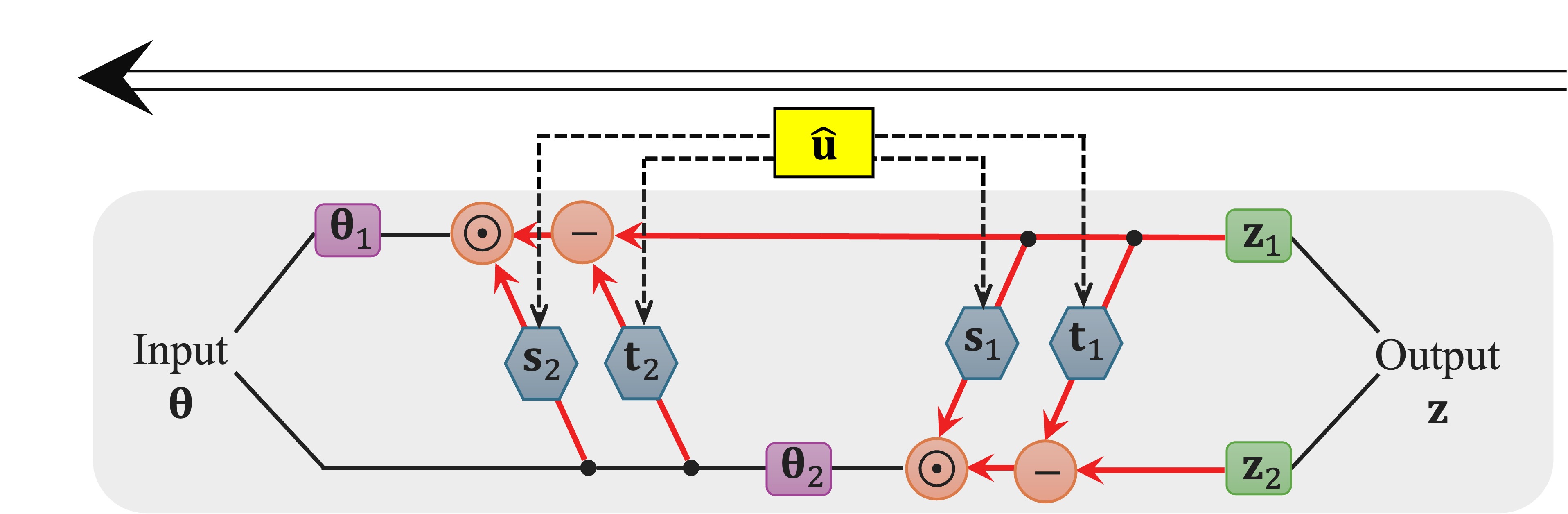}
        \caption{ inverse transformation }
        \label{fig:3b}
    \end{subfigure}
    \caption{Illustration of transformations in cACLs.}
    \label{fig:3}
\end{figure}

Dinh et al. \cite{dinh2016density} introduced ACLs as a technique to create invertible maps with cheap-to-evaluate Jacobian determinants.
Each ACL operates by partitioning the input vector \( \bm{\theta} \) into two segments, \( \bm{\theta}_1 \) and \( \bm{\theta}_2 \), and similarly dividing the output vector \( \mathbf{z} \) into \( \mathbf{z}_1 \) and \( \mathbf{z}_2 \). This division allows us to define invertible transformations in terms of embedded subnetworks, denoted as \( \mathbf{s}_1(\cdot) \), \( \mathbf{s}_2(\cdot) \), \( \mathbf{t}_1(\cdot) \), and \( \mathbf{t}_2(\cdot) \). These subnetworks, which can be chosen as arbitrary neural network architectures---like fully connected networks---are not required to be invertible. With the observed summary data \( \hat{\mathbf{u}} \) serving as an additional conditioning input, the forward transformations as shown in Fig. \ref{fig:3a} are formulated as
\begin{align}
\label{eq:25}
\mathbf{z}_1 &= \bm{\theta}_1 \odot \exp(\mathbf{s}_2(\bm{\theta}_2, \hat{\mathbf{u}})) + \mathbf{t}_2(\bm{\theta}_2, \hat{\mathbf{u}})\\
\label{eq:26}
\mathbf{z}_2 &= \bm{\theta}_2 \odot \exp(\mathbf{s}_1(\mathbf{z}_1, \hat{\mathbf{u}})) + \mathbf{t}_1(\mathbf{z}_1, \hat{\mathbf{u}})
\end{align}
where \( \odot \) symbolizes the element-wise product.

The forward transformation can be readily inverted. Given \( \mathbf{z} = (\mathbf{z}_1, \mathbf{z}_2) \), the inversion is accomplished by inversely transforming through cACLs. The inverse transformation, as shown in Fig. \ref{fig:3b}, is formulated as follows
\begin{align} 
\label{eq:27}
\bm{\theta}_2 = (\mathbf{z}_2 - \mathbf{t}_1(\mathbf{z}_1, \hat{\mathbf{u}})) \odot \exp(-\mathbf{s}_1(\mathbf{z}_1, \hat{\mathbf{u}})) \\
\label{eq:28}
\bm{\theta}_1 = (\mathbf{z}_1 - \mathbf{t}_2(\mathbf{\bm{\theta}}_2, \hat{\mathbf{u}})) \odot \exp(-\mathbf{s}_2(\mathbf{z}_2, \hat{\mathbf{u}})).
\end{align}
Eqs. \ref{eq:25}--\ref{eq:28} are essential in guaranteeing the tractability of the Jacobian determinant of the transformation, like in Eq. \ref{eq:objective}. Within an ACL, the Jacobian matrix is usually triangular, leading to cost-effective evaluation of its  determinant.

The Jacobian matrix can be written as

\begin{equation}\label{jacobian_matrix}
\frac{\partial (\bm{\theta}_1, \bm{\theta}_2)}{\partial (\mathbf{z}_1, \mathbf{z}_2)} =
\begin{pmatrix}
\frac{\partial \bm{\theta}_1}{\partial \mathbf{z}_1} & \frac{\partial \bm{\theta}_1}{\partial \mathbf{z}_2} \\
\frac{\partial \bm{\theta}_2}{\partial \mathbf{z}_1} & \frac{\partial \bm{\theta}_2}{\partial \mathbf{z}_2}
\end{pmatrix}
=
\begin{pmatrix}
\mathbf{I} & 0 \\
\frac{\partial \bm{\theta}_2}{\partial \mathbf{z}_1} & \frac{\partial \bm{\theta}_2}{\partial \mathbf{z}_2}.
\end{pmatrix}
\end{equation}
In this block triangular structure, the upper left block is the identity matrix \( \mathbf{I} \) because \( \bm{\theta}_1 \) is transformed independently of \( \mathbf{z}_2 \). The 0 in the top-right block reflects the fact that \( \bm{\theta}_1 \) is independent of \( \mathbf{z}_2 \).
The lower left block captures the dependence of \( \bm{\theta}_2 \) on \( \mathbf{z}_1 \), while the lower right block represents the transformation of \( \bm{\theta}_2 \) with respect to \( \mathbf{z}_2 \). The determinant of a block triangular matrix is simply the product of the determinants of the diagonal blocks. Because the upper left block is the identity matrix, its determinant is 1, and the overall determinant of the Jacobian matrix is determined by the diagonal terms.

\subsubsection{Conditional spline layers}

Rational-quadratic neural spline flows are designed to model complex distributions with high efficiency and flexibility \cite{durkan2019neural}.
These flow models employ monotonic rational-quadratic SLs as the fundamental building block for developing invertible functions. Each bin within these layers is characterized by a monotonically increasing rational-quadratic function. Rational-quadratic functions, defined as the ratio of two quadratic polynomials, are not only differentiable but also analytically invertible when restricted to monotonic segments. The knot points refer to the boundaries between consecutive bins in the spline. At each knot point, the spline must satisfy certain conditions, such as continuity in both the function's value and its derivative, ensuring smooth transitions between the segments. SLs provides more flexibility than traditional quadratic functions because they allow direct control over the values and derivatives at each knot point.
A monotonic rational-quadratic spline is designed to map an interval of \([-B,B]\) onto itself. Outside of this range, the transformation defaults to an identity function, thereby introducing linear ``tails'' to the curve.
\begin{figure}[ht]
    \centering
    \includegraphics[width=0.5\textwidth]{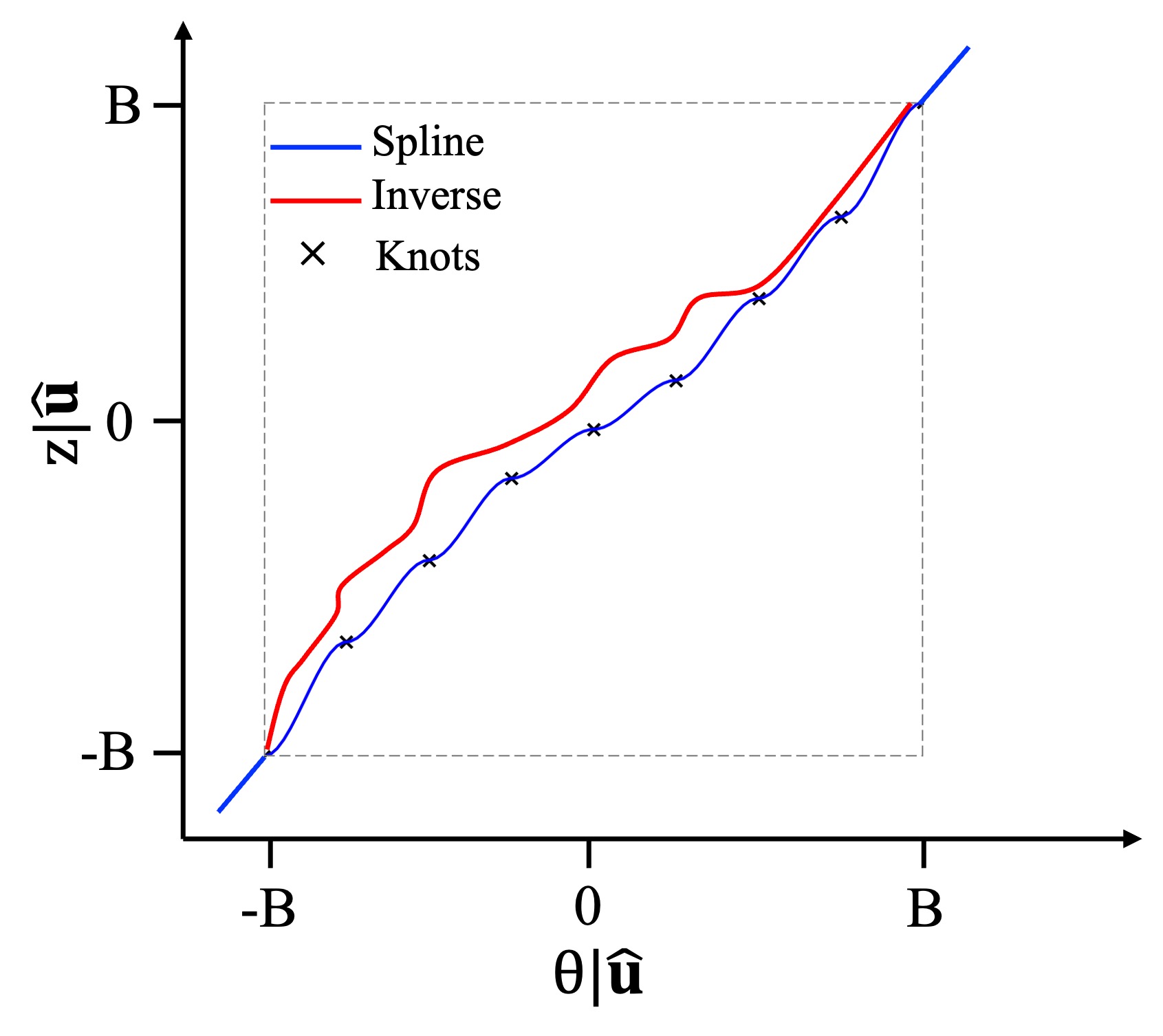} 
    \caption{Monotonic rational-quadratic transformation (spline layer).}
    \label{fig:5}
\end{figure}

Fig. \ref{fig:5} illustrates a monotonic rational-quadratic transformation, also known as a spline layer,
which is conditioned on the observed data \( \hat{\mathbf{u}} \).
Splines are applied component-wise,
meaning that each dimension is transformed independently, avoiding the exponential growth in the number of knots.
This ensures computational feasibility as the number of dimensions increases.
In the one-dimensional case, there are \( K + 1 \) knots defining the intervals between consecutive segments of the spline. For the remainder of this section, we will denote the $i$th component of $\bm{\theta}$ and $\mathbf{z}$ as $\theta$ and $z$, respectively.
This spline is composed of segments that lie between \( K+1 \) coordinate pairs known as knots \( (\theta_k, z_k) \), for \( k = 0, \ldots, K \).
These knots are organized in a sequence that increases monotonically, starting from \( (\theta_0, z_0) = (-B, -B) \) and ending at \( (\theta_K, z_K) = (B, B) \).
To define the slope at the internal knots of the spline, \( K-1 \) arbitrary positive values are assigned and denoted as \( \tau^{(k)} \), where \(k\) ranges from 1 to \(K-1\). The derivatives at the boundaries are set to 1,\( \tau^{(0)} \) and \( \tau^{(K)} \) are set to one, ensuring the slopes at the endpoints are consistent with the linear tails of the spline.

Let \( \phi_k \) be the slope of the segment between two consecutive knots: \( \phi_k = \frac{{z}_{k+1} - {z}_k}{{\theta}_{k+1} - {\theta}_k} \). Additionally, define \( \xi({\theta}) \) as the normalized position of \( {\theta} \) within the \( k \)th interval: \( \xi({\theta}) = \frac{{\theta} - {\theta}_k}{{\theta}_{k+1} - {\theta}_k} \). In SLs, the parameters of the rational-quadratic spline (such as the widths and heights of the bins, as well as the derivatives at the knots) are predicted by a neural network. The neural network outputs the parameters that control the spline transformation. Specifically, the networks outputs the widths, heights, and slopes at the internal knots, ensuring that the transformation is invertible and smooth across the bins.

Using these definitions, the transformation \( \mathbf{h}({\theta}) \) in the \( k \)th bin of the spline is given by the ratio of two functions, \( \omega_1^{(k)}(\xi) \) and \( \omega_2^{(k)}(\xi) \), expressed as:
\begin{equation} \label{eq:31}
\frac{\omega_1^{(k)}(\xi)}{\omega_2^{(k)}(\xi)} = {z}^{(k)} + \frac{({z}^{(k+1)} - {z}^{(k)}) \left[ {\phi}_k \xi^2 + \tau^{(k)} \xi (1 - \xi) \right]}{{\phi}_k + (\tau^{(k+1)} + \tau^{(k)} - 2{\phi}_k \xi)\xi (1 - \xi)}
\end{equation}
where \( \omega_1^{(k)}(\xi) \) and \( \omega_2^{(k)}(\xi) \) denote the quadratic polynomials governing the transformation within the \( k \)th bin. The parameters of these polynomials, including the slopes and bin heights and width, are determined by the neural network. The rational-quadratic spline is constructed within specified bounds \( (-B, B) \times (-B, B) \), with the identity function applied outside this range to ensure numerical stability. The value of \( B \) is typically chosen before training to cover the expected range of the input \( {\theta} \), ensuring that the transformation remains well-behaved across the entire domain

Given the element-wise and monotonic properties of the rational-quadratic transformation, it is feasible to efficiently compute the logarithm of the absolute value of the determinant of the Jacobian matrix of the transformation. This computation involves summing the logarithms of the derivatives of the function with respect to \( {\theta} \). The inverse of the rational-quadratic function, such as \( \xi({\theta}) \), can be analytically determined by inverting Eq. \ref{eq:31}, which is equivalent to finding the roots of a quadratic equation \cite{durkan2019neural}.

\begin{figure}[ht]
    \centering
    \includegraphics[width=\textwidth]{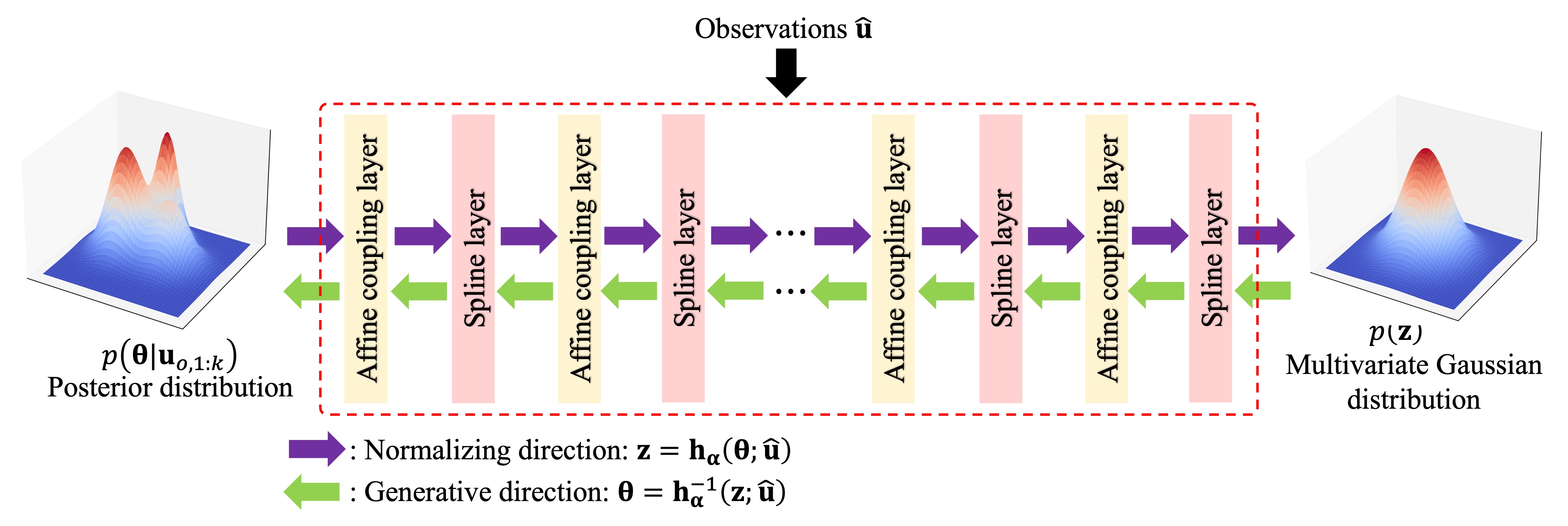} 
    \caption{The proposed conditional inference network.}
    \label{conditionalnetwork}
\end{figure}

Finally, we put cACLs and cSLs together to build the architecture of our proposed inference network, which is conditional on summary features \( \hat{\mathbf{u}} \), as shown in Fig. \ref{conditionalnetwork}. Our design strategy employs a composition of multiple cACLs and cSLs arranged sequentially to approximate a complex posterior. This sequence of layers collectively acts as a deep neural network, with the network parameters denoted by \( \bm{\alpha} \).
The network structure is characterized by an alternation between cACLs and cSLs, ensuring the necessary depth and complexity for estimating complex posteriors.
At each stage of the inference process, the output of a given cACL is channeled as the input to the subsequent cSL, thus creating a chain of transformations.
In summary, the inference process is bidirectional: the normalizing direction takes the parameters \( \bm{\theta} \) and the observations \( \hat{\mathbf{u}} \) to map them to a latent Gaussian variable \( \mathbf{z} \), such as \( \mathbf{z} = \mathbf{h}_{\bm{\alpha}}(\bm{\theta}; \hat{\mathbf{u}}) \). Conversely, the generative direction recovers the parameters of interest from the latent space through the inverse function \( \bm{\theta} = \mathbf{h}_{\bm{\alpha}}^{-1}(\mathbf{z}; \hat{\mathbf{u}}) \).

\subsection{Amortized inference}

\begin{figure}[ht]
    \centering
    \includegraphics[width=\textwidth]{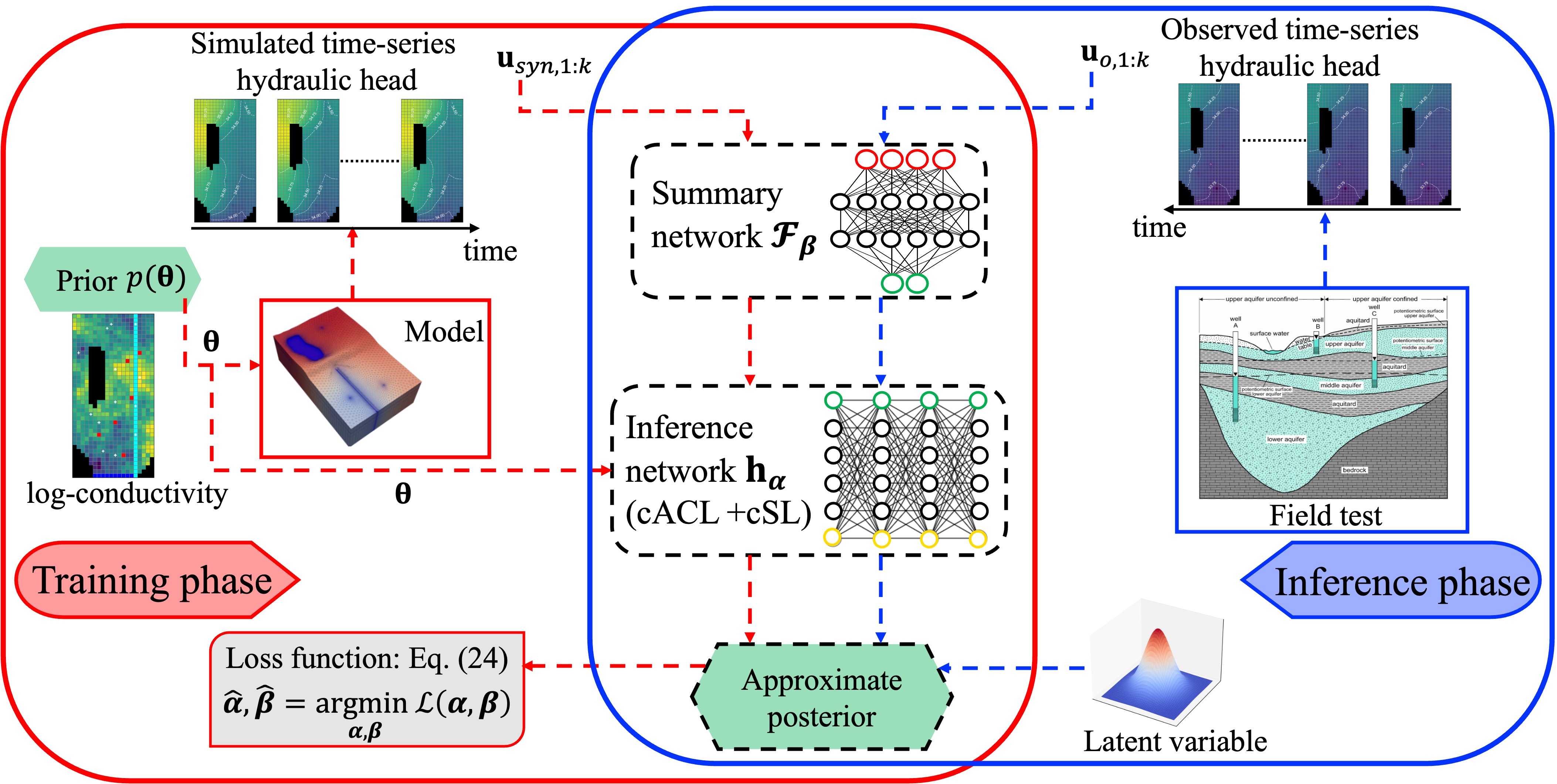} 
    \caption{The workflow of proposed amortized likelihood-free inference.}
    \label{summary}
\end{figure}

Unlike traditional Bayesian inference methods, which necessitate repeating the entire inference procedures for each new dataset, our proposed likelihood-free approach amortizes the workflow by separating it into a computationally demanding training phase followed by a more cost-efficient inference phase. Fig. \ref{summary} depicts the schematic of our proposed method. First, we generate synthetic training data using a forward model and a prior distribution.
During training, the synthetic data \( \{ \bm{\theta}^{(i)}, \mathbf{u}^{(i)}_{\text{syn},1:k} \} \), is first processed
in batches by the summary network \( \mathcal{\bm{F}}_{\bm{\beta}} \), producing the set of summary features \( \{ \hat{\mathbf{u}}^{(i)} \}\) that are then fed into the inference network. The networks are jointly optimized by minimizing the loss function in Eq. \ref{eq:objective_joint_final} with respect to all network parameters. During the inference phase, observed data \( \mathbf{u}_{o,1:k} \), is input into the trained networks. The posterior distribution is approximated by drawing samples from the latent distribution and leveraging the inference network.

\subsection{Application} \label{sec:application}

We demonstrate our method
by applying it to a parameter estimation problem for the synthetic groundwater model proposed in \cite{freyberg1988exercise}. The MODFLOW implementation of this model is described in \cite{wang2024bayesian}. 
The model domain consists of a single layer with 40 rows and 20 columns, where each cell is a square with a side length of 250 meters, resulting in a total domain extent of 10,000 meters by 5,000 meters. We compare the proposed method against a likelihood-based benchmark method, namely the PEST implementation of the iterative ensemble Kalman smoother \cite{white2018model}, PEST-IES.

\begin{figure}[ht]
    \centering
    \includegraphics[width=\textwidth]{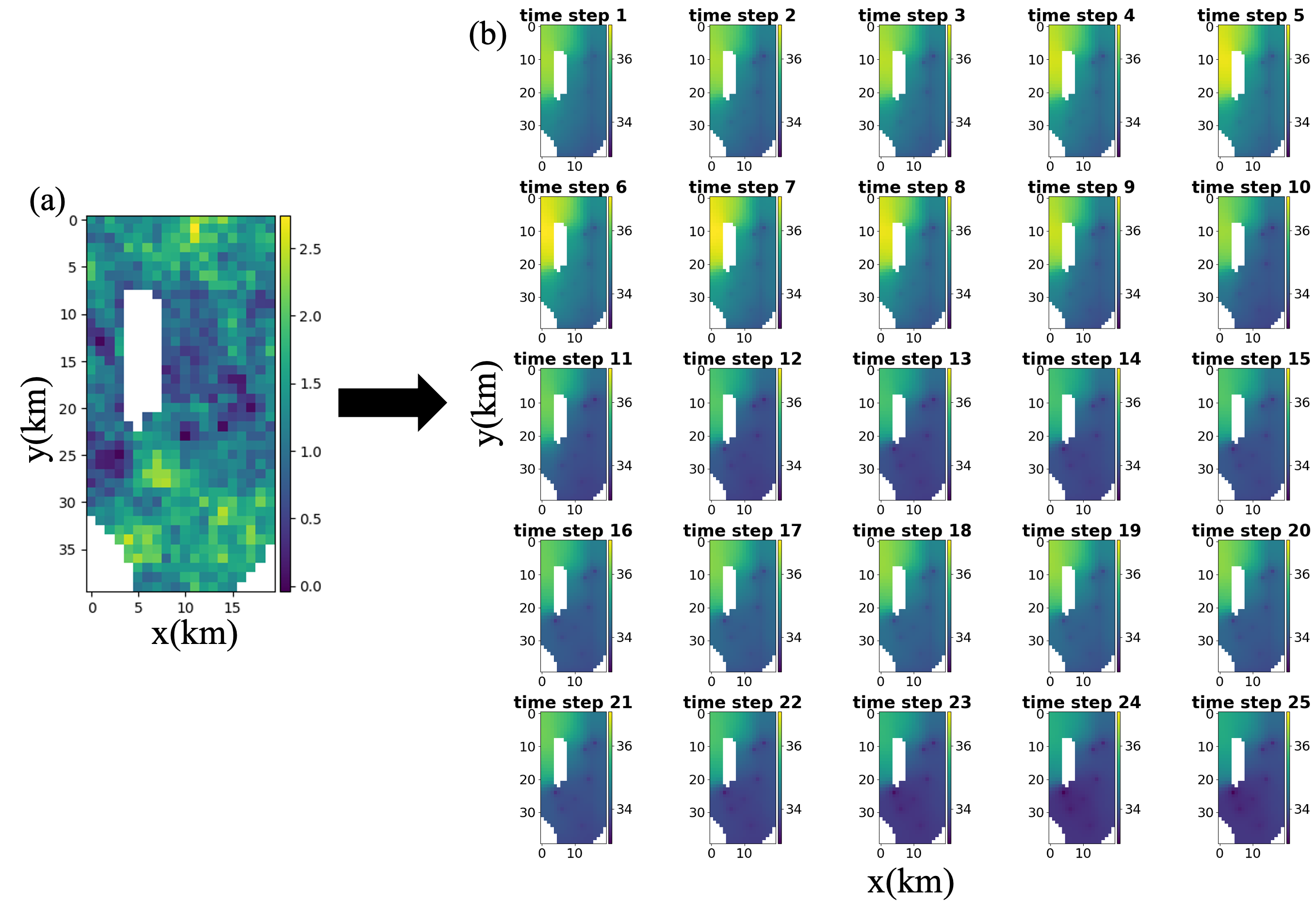} 
    \caption{A realization synthetic data at 706 locations: (a) log-conductivity field; (b) hydraulic head field.}
    \label{fig:8}
\end{figure}

As described in Section~\ref{sec:inverse formulation}, our case study focuses on estimating the hydraulic log-conductivity field of the 706 control volumes. The observed data consists of synthetic, noisy observations of the hydraulic head at 13 observation locations over 25 timesteps.
The log-conductivity prior distribution is a Gaussian process prior with zero mean and the exponential covariance function. 
From this prior, 5,001 synthetic datasets are generated using \textsc{MODFLOW}. Fig.~\ref{fig:8} illustrates a single realization of this synthetic data, showing the input conductivity field alongside the hydraulic head field across the simulation domain at the 25 timesteps. Cells colored white are inactive, meaning the governing equation is not solved in these cells, while the remaining cells are active.
Additionally, to simulate measurement errors, random Gaussian noise with zero mean and standard deviation of 0.01 is added to the synthetic observations.

\subsubsection{Model training}

We partitioned the 5,001 datasets into 4,800 for training, 100 for validation, and the remainder for testing. The summary and inference networks are implemented in TensorFlow \cite{tensorflow2015-whitepaper} and trained by minimizing the loss function Eq. \ref{eq:objective_joint_final} using the Adam optimizer \cite{KingBa15}. The learning rate is initially set to $1 \times 10^{-3}$ and reduced with a decay rate of 0.95. The inference network's architecture comprises five cACLs, each with four fully connected neural networks, and five cSLs, each containing 16 bins. The cACLs and cSLs are alternately stacked. A dropout rate of 0.5 is implemented at each cACL and cSL
to prevent overfitting. The summary network is structured as a 1D convolutional neural network, compressing observation data from a $25\times13$ matrix to a $1\times256$ vector. Training was conducted over 4,000 epochs, with each epoch comprising batches of 120 datasets.

\begin{figure}[ht]
  \centering
  \includegraphics[width=\textwidth]{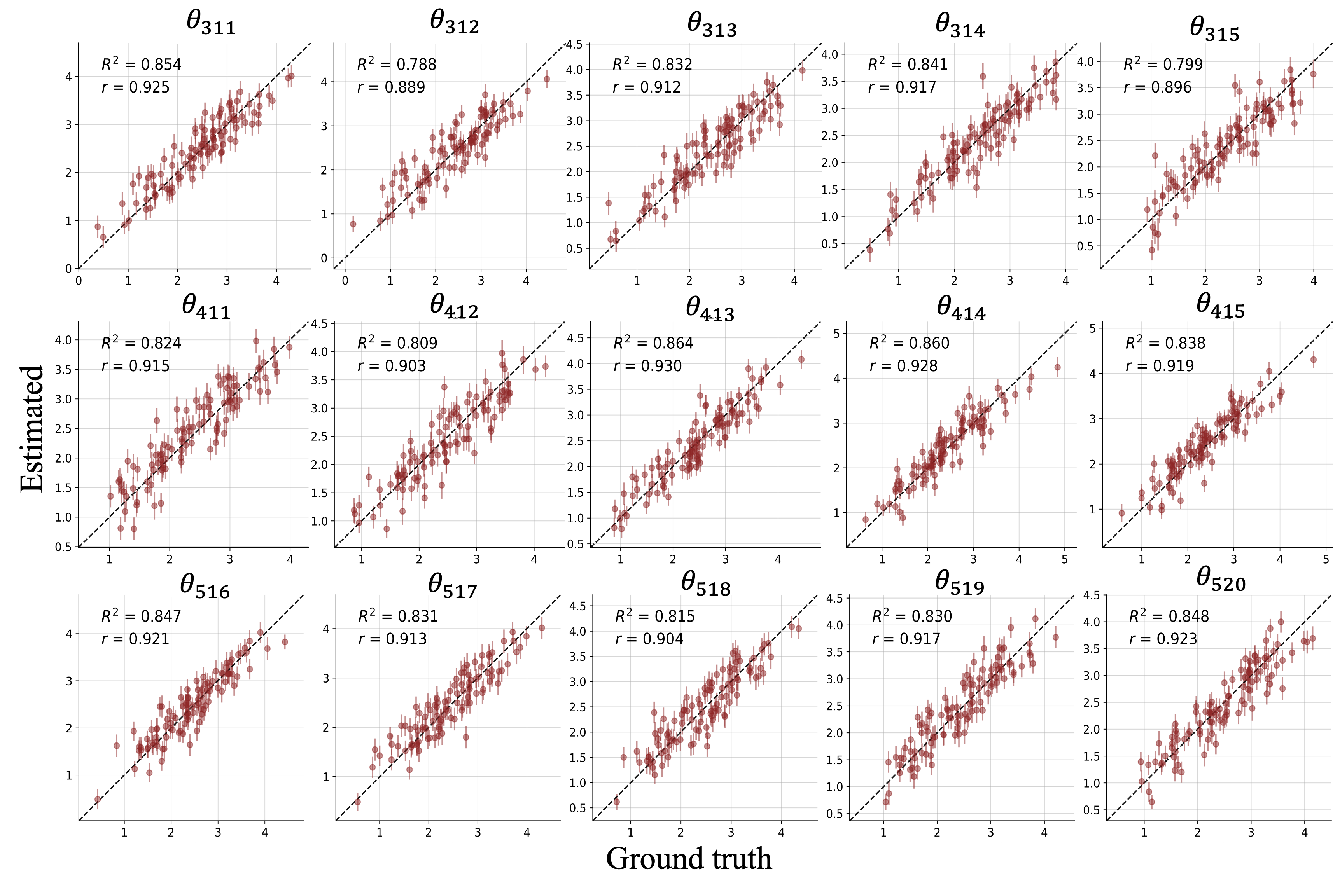}
  \caption{Validation of trained model performance.}
  \label{fig:9}
\end{figure}

Fig. \ref{fig:9} shows the performance of the trained model
for 100 test datasets. Due to space limitations in the paper, we show results for only a subset of the 706 parameters. The error bars depict the \(95\%\) credibility
intervals, calculated from 2,000 posterior samples. The results show that the estimated parameters generally align well with the ground truth, as evidenced by 87\% (13 out of 15) of the selected parameters achieving a coefficient of determination \(R^2\) above 0.8.
Furthermore, the correlation coefficient \((r)\) confirms a strong linear relationship between the estimated posterior mean and actual values.
However, it is noted that certain parameters, such as \(\theta_{312}\) and \(\theta_{315}\), exhibit lower \(R^2\) values compared to others. This variation is attributed to their relative insensitivity to the observations collected from the 13 sensor locations. The 15 parameters shown were selected randomly for illustrative purposes.

\subsubsection{Results of parameter estimation}
  
We compare our method against PEST-IES for two test problems corresponding to two reference conductivity fields.
It is important to highlight the workflow differences between the methods: PEST-IES is likelihood-based and thus relies on a forward model to estimate parameters, whereas our proposed method bypasses the evaluation of the likelihood function. We employed three metrics to compare the accuracy of the parameter estimates derived from both methods:
\begin{itemize}
  \item To quantify the precision of each estimated parameter, we use the relative error, defined as
  \[
  \text{relative error}_i = \frac{|\theta_{i, \text{est}} - \theta_{i, \text{ref}}|}{|\theta_{i, \text{ref}}|}, \quad i = 1,2,\ldots,706,
  \]
  where \(\theta_{i, \text{est}}\) and \(\theta_{i, \text{ref}}\) are the \(i\)th estimated and true parameters, respectively. A lower relative error signifies higher accuracy in the parameter estimation.

  \item For assessing the overall accuracy, we employ the relative \(\ell_2\) error, defined by the formula
  \[
  \ell_2\text{ error} = \frac{\sqrt{\sum_{i=1}^{706} (\theta_{i, \text{est}} - \theta_{i, \text{ref}})^2}}{\sqrt{\sum_{i=1}^{706} (\theta_{i, \text{ref}})^2}}.
  \]
  A smaller relative \(\ell_2\) error indicates more precise estimation of parameters across the model.

  \item The log predictive probability (\(\text{LPP}\)) is another metric used, calculated by
  \[
  \text{LPP} = \sum_{i=1}^N \left\{\frac{(\theta_{i, \text{est}} - \theta_{i, \text{ref}})^2}{2(\theta_{i, \text{ref}})^2} + \frac{1}{2} \log(2\pi(\theta_{i, \text{ref}})^2)\right\}.
  \]
  A higher \(\text{LPP}\) value points to a more accurate model estimation because it reflects a better fit of the estimated parameters to the true values.
\end{itemize}

\begin{figure}[ht]
    \centering
    \includegraphics[width=0.9\textwidth]{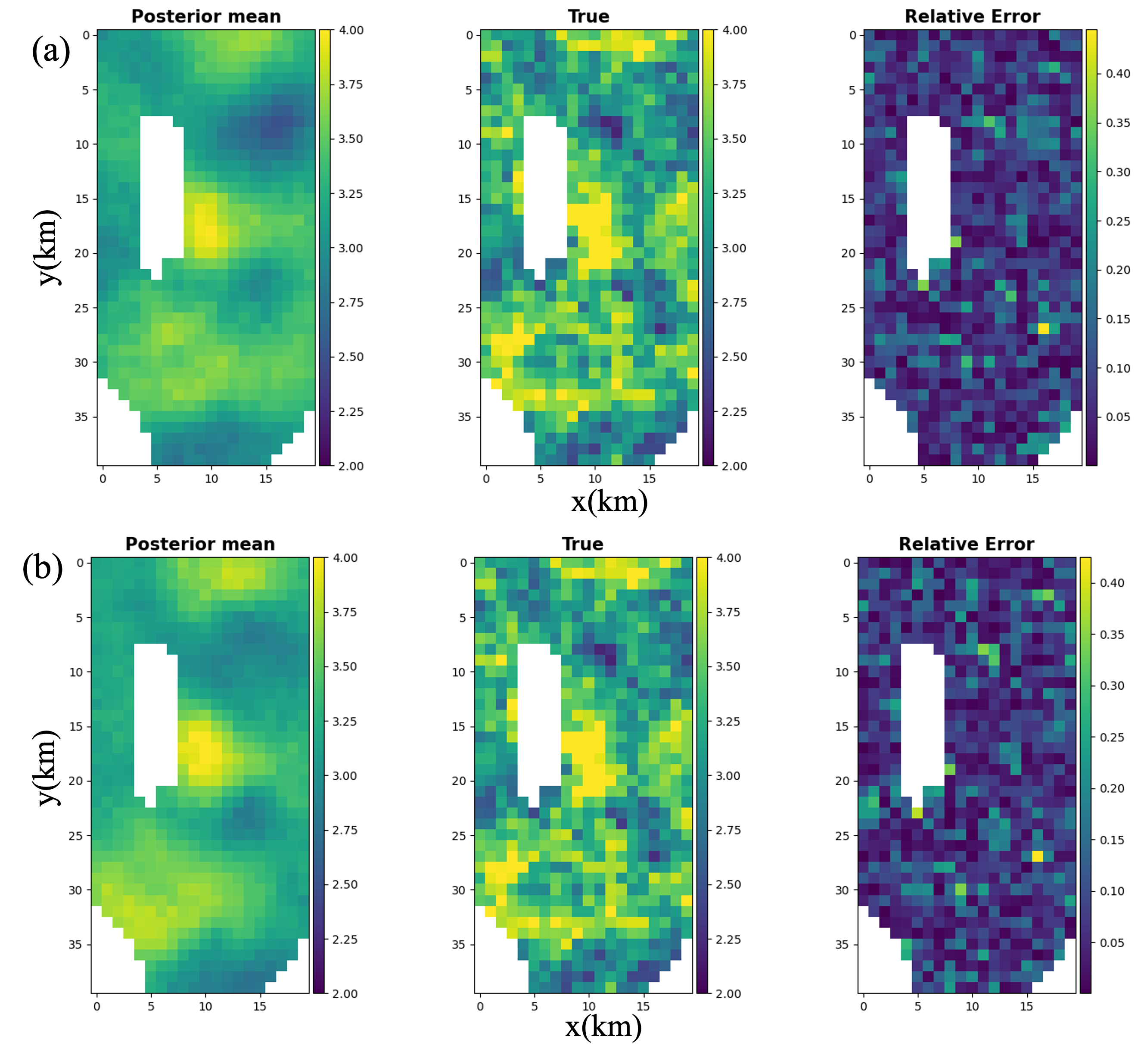} 
    \caption{Estimation of log-conductivity field using test \#1: (a) PEST-IES, (b) proposed.}
    \label{fig:10}
\end{figure}

\begin{figure}[ht]
    \centering
    \includegraphics[width=0.9\textwidth]{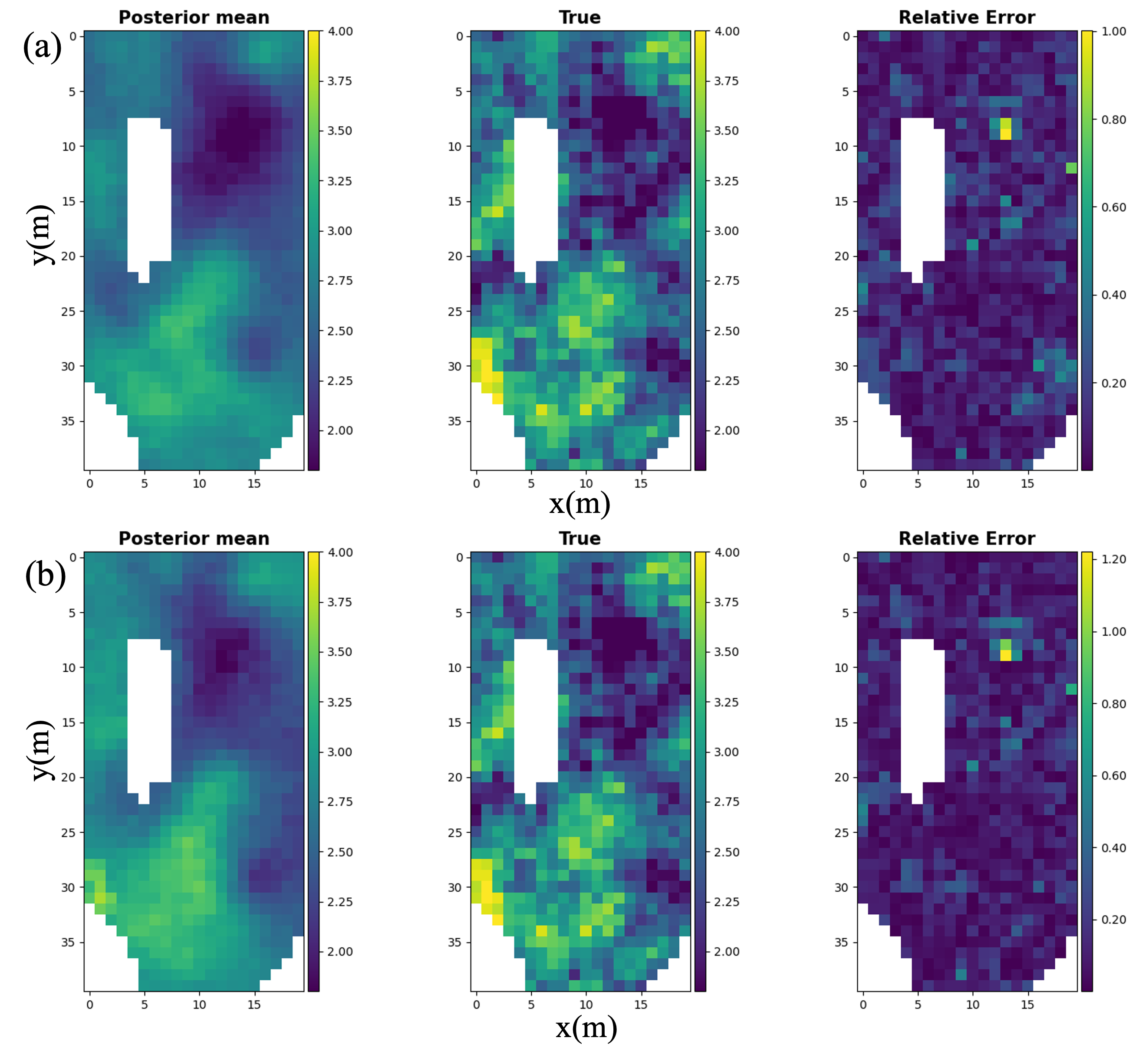} 
    \caption{Estimation of log-conductivity field using test \#2: (a) PEST-IES, (b) proposed.}
    \label{fig:11}
\end{figure}

\begin{figure}[ht]
    \centering
    \includegraphics[width=\textwidth]{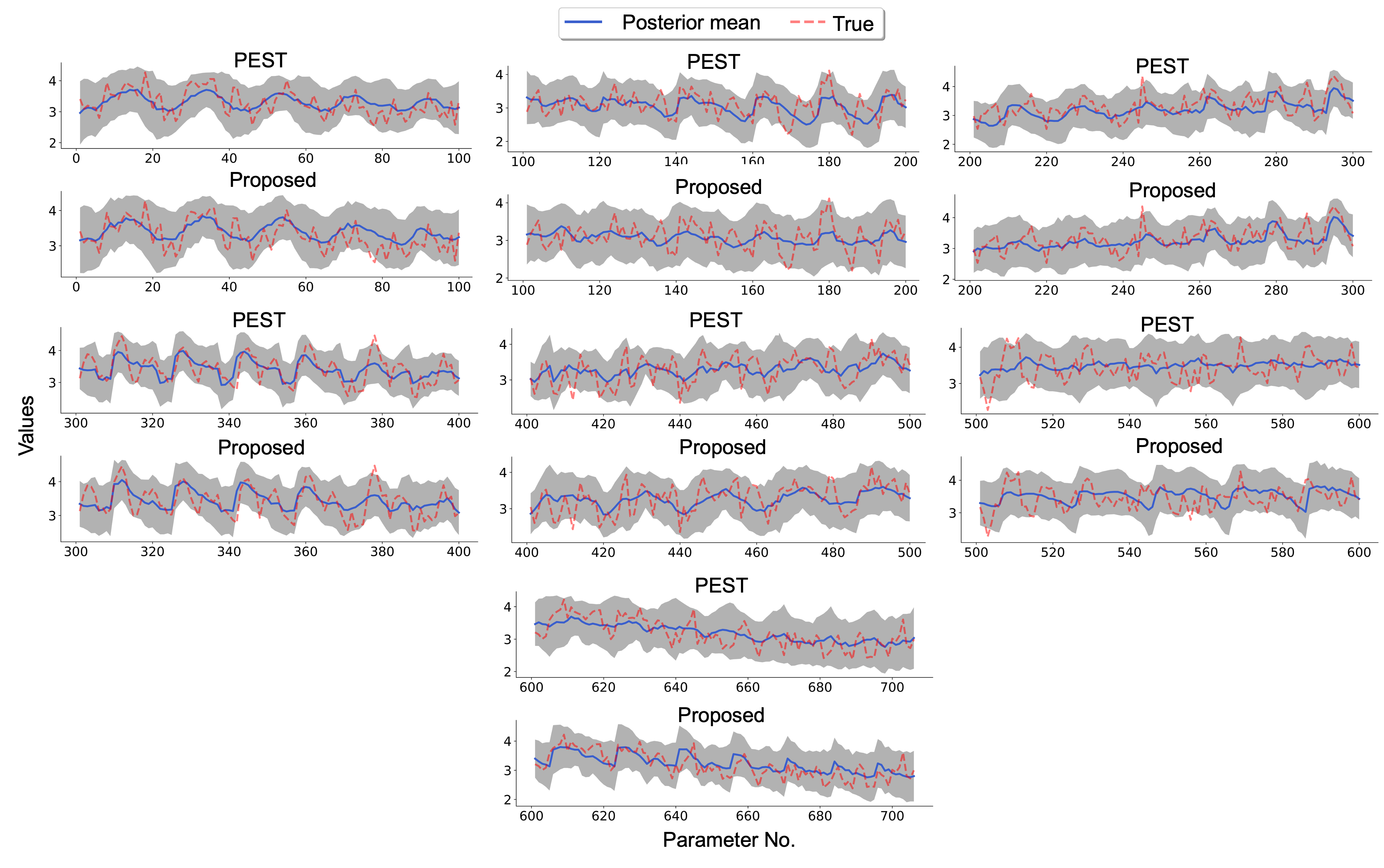} 
    \caption{Estimation of 706 parameters using test \#1.}
    \label{fig:12}
\end{figure}

\begin{figure}[ht]
    \centering
    \includegraphics[width=\textwidth]{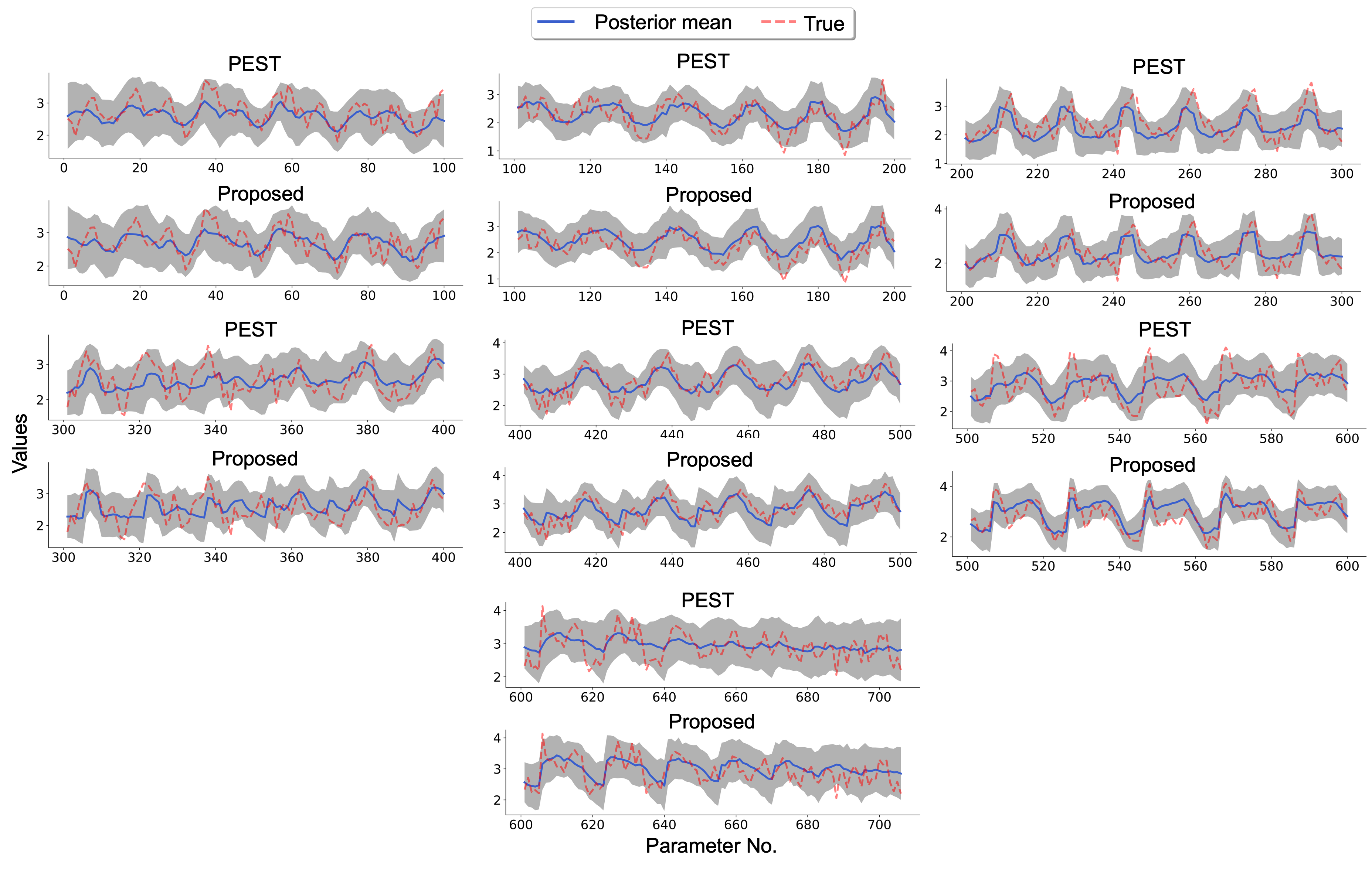} 
    \caption{Estimation of 706 parameters using test \#2.}
    \label{fig:13}
\end{figure}

Figs. \ref{fig:10} and \ref{fig:11} display the posterior mean log-conductivity fields estimated with PEST-IT and the likelihood-free method, true log-conductivity fields, and point relative errors for tests \#1 and \#2, respectively.
It can be seen that both methods yield comparably accurate estimations, which suggests that the proposed method can reliably reconstruct the conductivity field without the need for likelihood function evaluation.
Figs. \ref{fig:12} and \ref{fig:13} show the true parameter values, the posterior mean values, and the corresponding \(95\%\) credibility intervals (shown in grey shaded areas) for both PEST-IES and the proposed method.
A majority of the estimated parameters are contained within the credibility interval, which suggests that the proposed method is reliable and yields estimations that are comparable to those obtained via the PEST-IES benchmark method. Furthermore, these figures highlight that parameter estimation for test \#2 poses a greater challenge, evidenced by the larger variability among the 706 parameters, particularly in the 200--300 and 500--600 location ranges, in comparison to test \#1. Despite these challenges, the proposed method manages to produce accurate parameter estimations, demonstrating its efficacy even under challenging conditions.

Table \ref{tab1} presents the accuracy metrics for the proposed PEST-IES methods for two test problems. For test \#1, both methods exhibit comparable accuracy, as measured by the relative \(\ell_2\) error and the LPP, with the proposed method showing a marginally lower average relative error compared to PEST-IES. When evaluating test \#2, the proposed method demonstrates slightly better performance, with a lower average relative error, a lower relative \(\ell_2\) error, and a marginally higher LPP score.
While these differences are modest, they suggest that the proposed method maintains its precision in parameter estimation, even with the increased variability among the 706 parameters in test \#2. Moreover, while the proposed and PEST-IES methods show similar performance levels in parameter estimation, the computational efficiency between the two is noteworthy (details are presented in Sec. \ref{sec:cost}). The PEST-IES method requires repeated evaluations of the forward model, which is computationally intensive. In contrast, the proposed method does not require the forward model evaluations outside the training data generation stage.

\begin{table}
\centering
\caption{Comparison between proposed and PEST-IES method.}
\begin{tabular}{@{}ccccccc@{}}
\toprule
\multirow{2}{*}{Method} & \multicolumn{3}{c}{Test \#1} & \multicolumn{3}{c}{Test \#2} \\
\cmidrule(r){2-4} \cmidrule(l){5-7}
          & \makecell{Average relative\\error} & \makecell{Relative \(\ell_2\)\\error} & {LPP} & \makecell{Average relative\\error} & \makecell{Relative \(\ell_2\)\\error} & {LPP} \\
\midrule
Proposed  & 8.13\% & 10.02\% & -229.44 & 11.62\% & 13.14\% & -271.29 \\
\addlinespace
PEST-IES      & 8.41\% & 10.08\% & -229.93 & 11.89\% & 13.85\% & -295.08 \\
\bottomrule
\end{tabular}
\label{tab1}
\end{table}

\begin{figure}[ht]
    \centering
    \includegraphics[width=\textwidth]{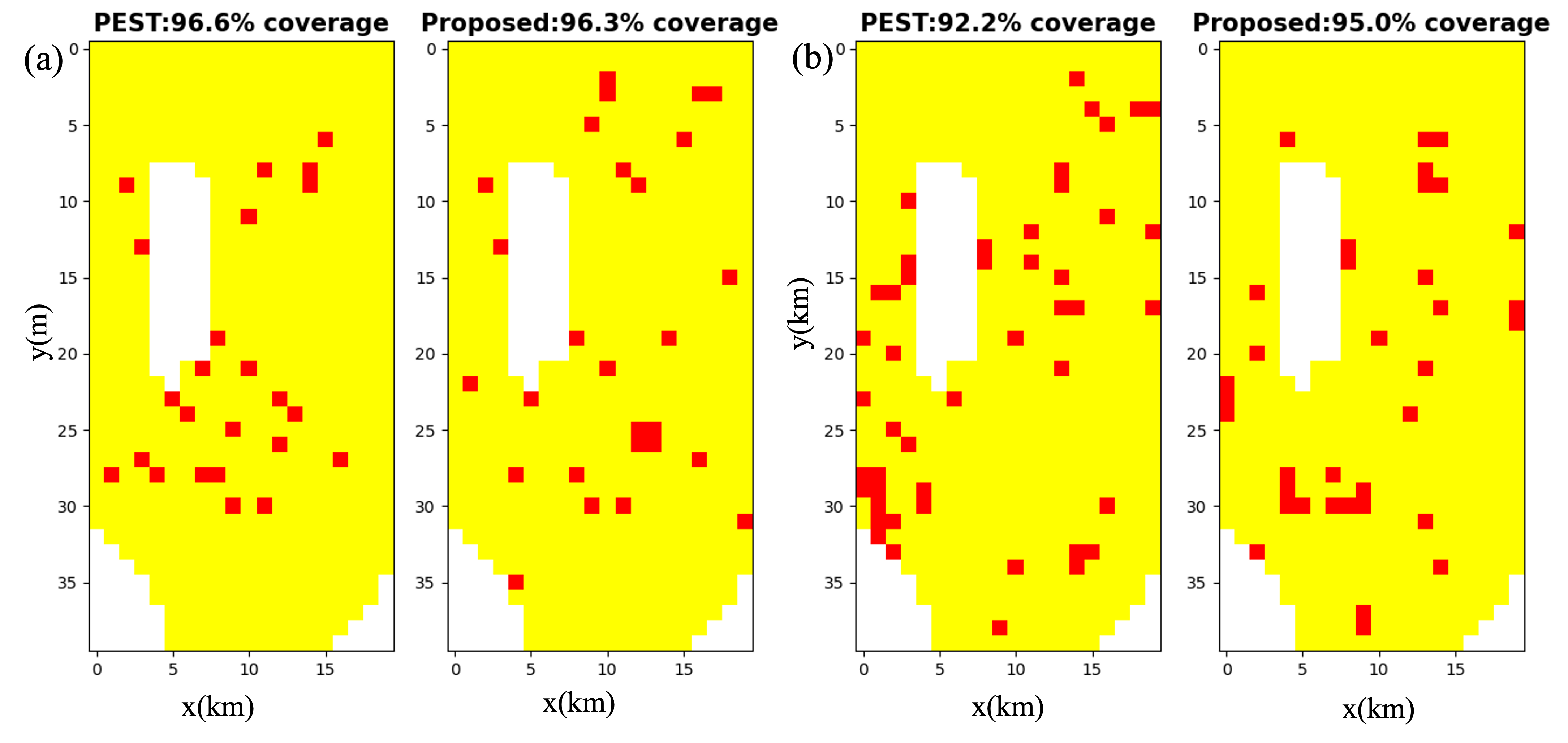} 
    \caption{Parameter coverage: (a) test \#1; (b) test \#2.}
    \label{fig:14}
\end{figure}

Fig. \ref{fig:14} shows the coverage map for each method and each of the two datasets under consideration. In this figure, yellow control volumes indicate that the true value is within the \(95\%\) credibility interval, and a red control volume indicates that the true parameter is outside said interval. For test \#1, PEST-IES and the proposed methods demonstrate similar coverages, indicating comparable reliability. However, for test \#2, which presents a more complex challenge for parameter estimation, both methods yield lower coverage compared to test \#1. Despite this, the proposed method achieves a higher coverage value closer to the expected value of \( 95\% \).

\begin{figure}[ht]
    \centering
    \includegraphics[width=\textwidth]{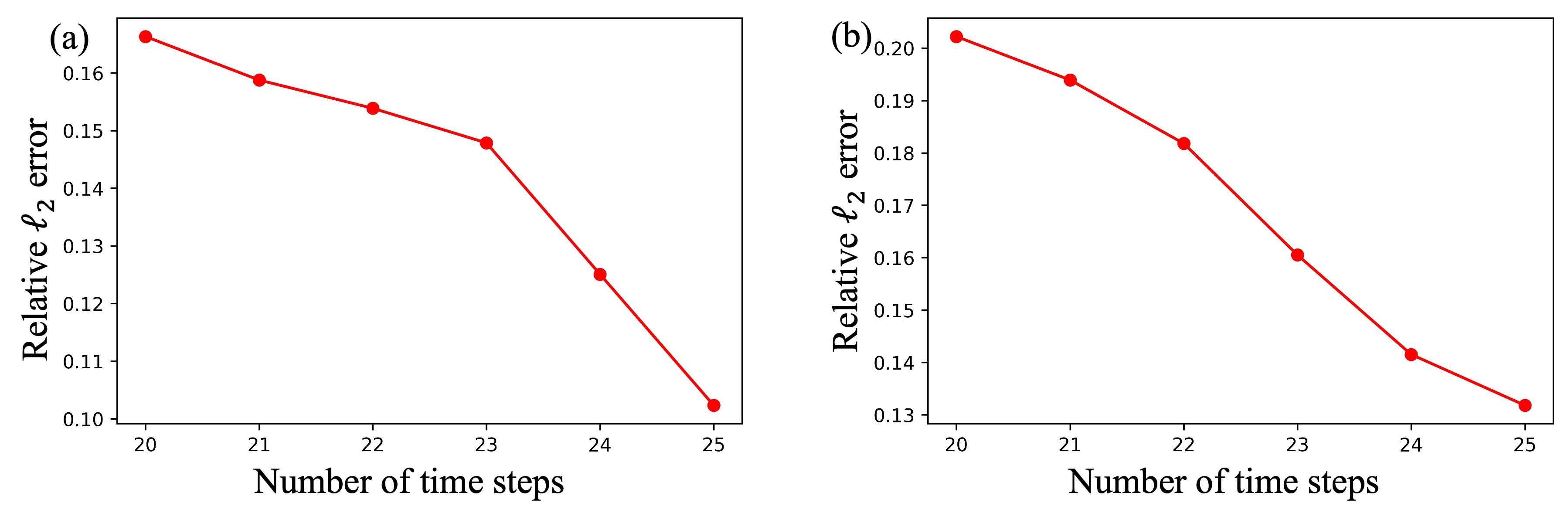} 
    \caption{Parameter estimation by varied data length: (a) dataset \#1; (b) dataset \#2.}
    \label{variedlength}
\end{figure}

Some likelihood-free approaches, such as ABC, and likelihood-based methods, including PEST-IES, require resampling or solving minimization problems for each new set of measurements. 
One of the attractive attributes of the proposed method is that it does not have to be retrained when measurements change. It can perform parameter estimation across different measurement set sizes utilizing a single trained model. To illustrate this adaptability, we considered five scenarios where the number of measurements at each location varies from 20 to 25 with a time increment of one year between measurements. In the training dataset, the number of observations at each location is 25.  Fig. \ref{variedlength} shows the relative \(\ell_2\) errors for various numbers of measurements. As anticipated, the relative \(\ell_2\) errors for both test problems generally decrease as the number of observations increases. Importantly, the findings in Fig. \ref{variedlength} confirm that the proposed model, once trained, can estimate parameters from datasets of arbitrary sizes. 

\subsubsection{Observation prediction}

\begin{figure}[ht!]
  \centering
  
  \begin{subfigure}[b]{\textwidth}
    \includegraphics[width=\textwidth]{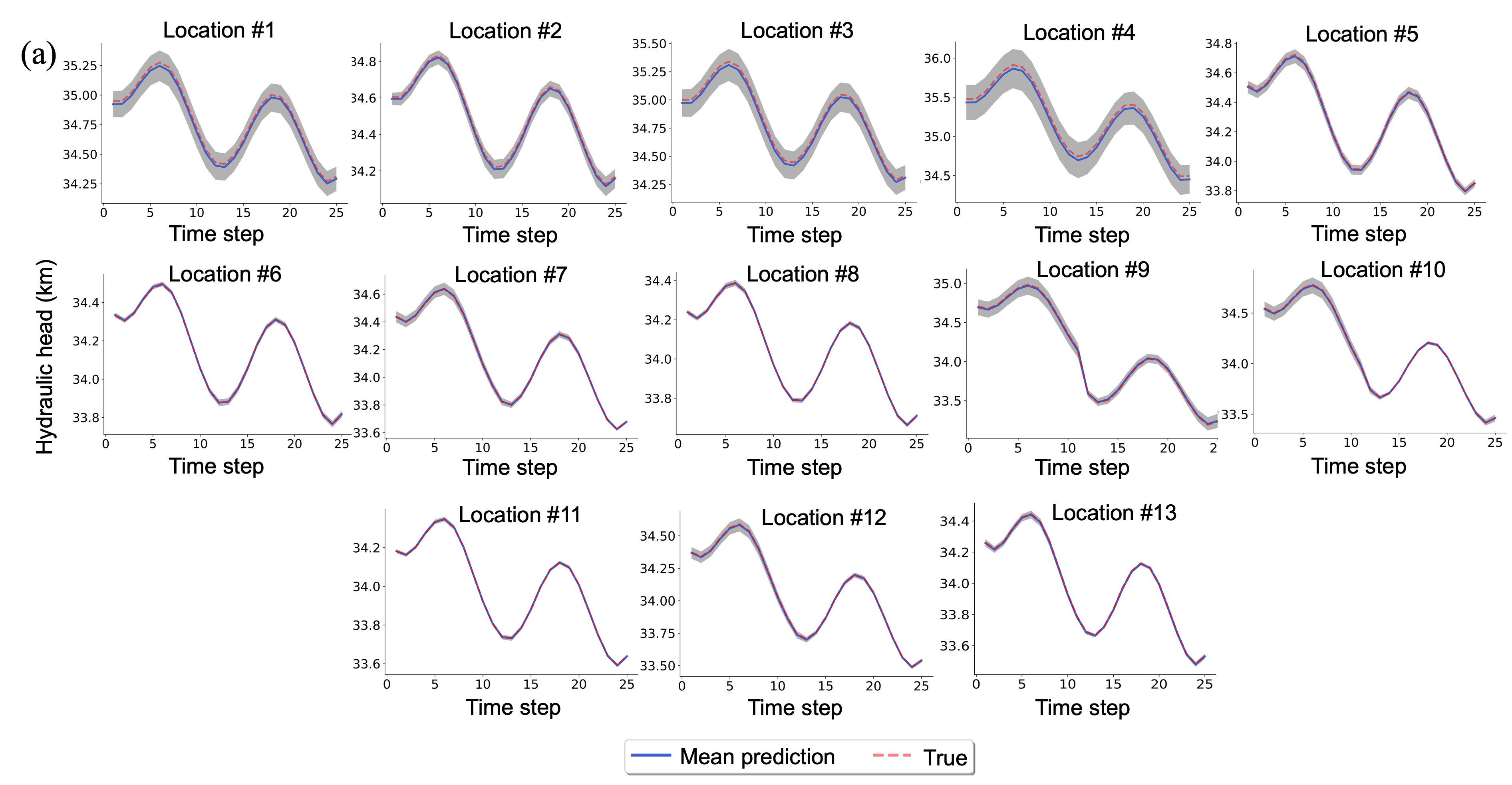}
  \end{subfigure}
  \begin{subfigure}[b]{\textwidth}
    \includegraphics[width=\textwidth]{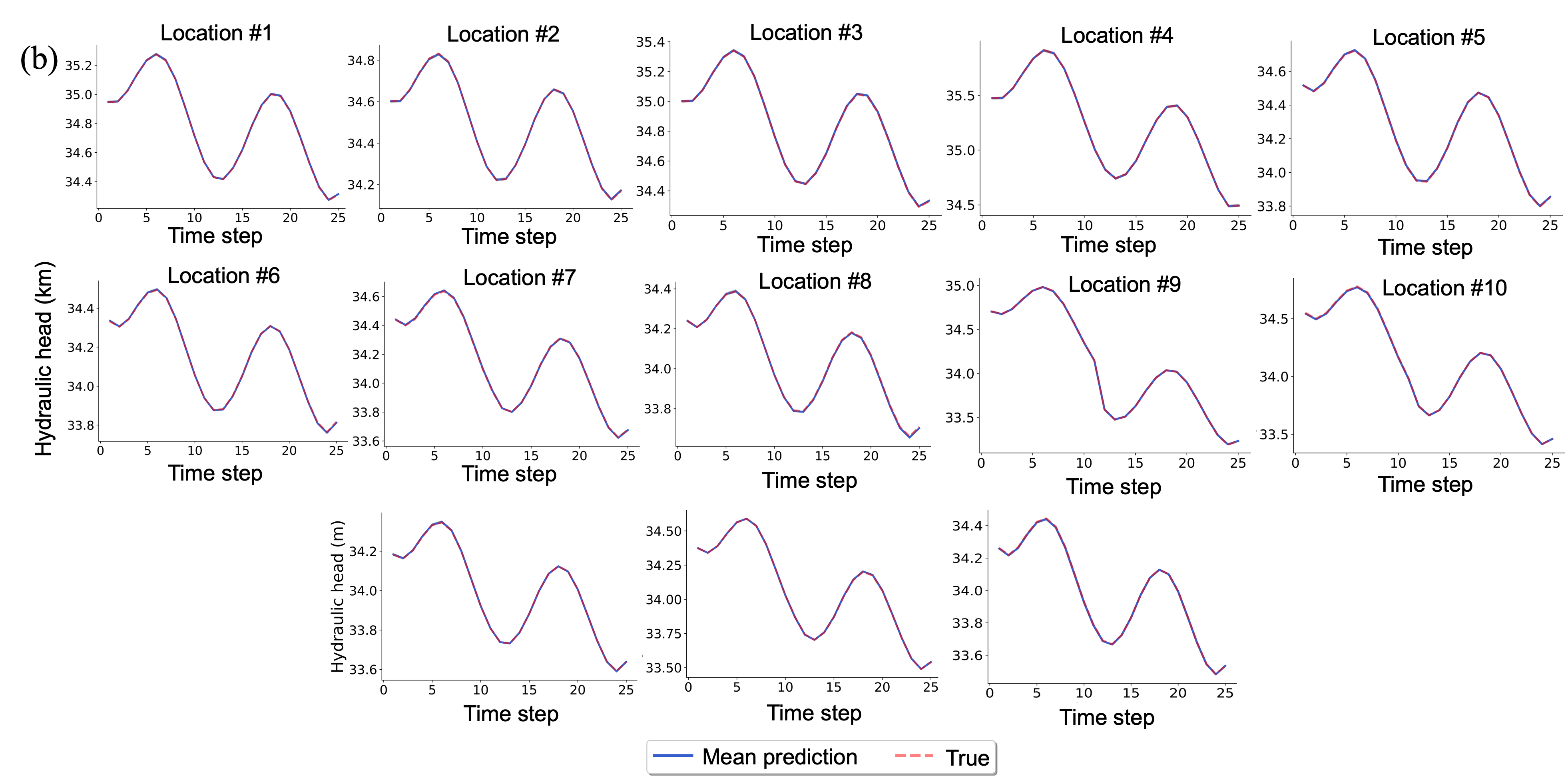}
  \end{subfigure}
  
  \caption{Observation prediction at 13 locations for test \#1: (a) proposed method; (b) PEST-IES.}
  \label{fig:16}
\end{figure}

\begin{figure}[ht!]
  \centering
  
  \begin{subfigure}[b]{\textwidth}
    \includegraphics[width=\textwidth]{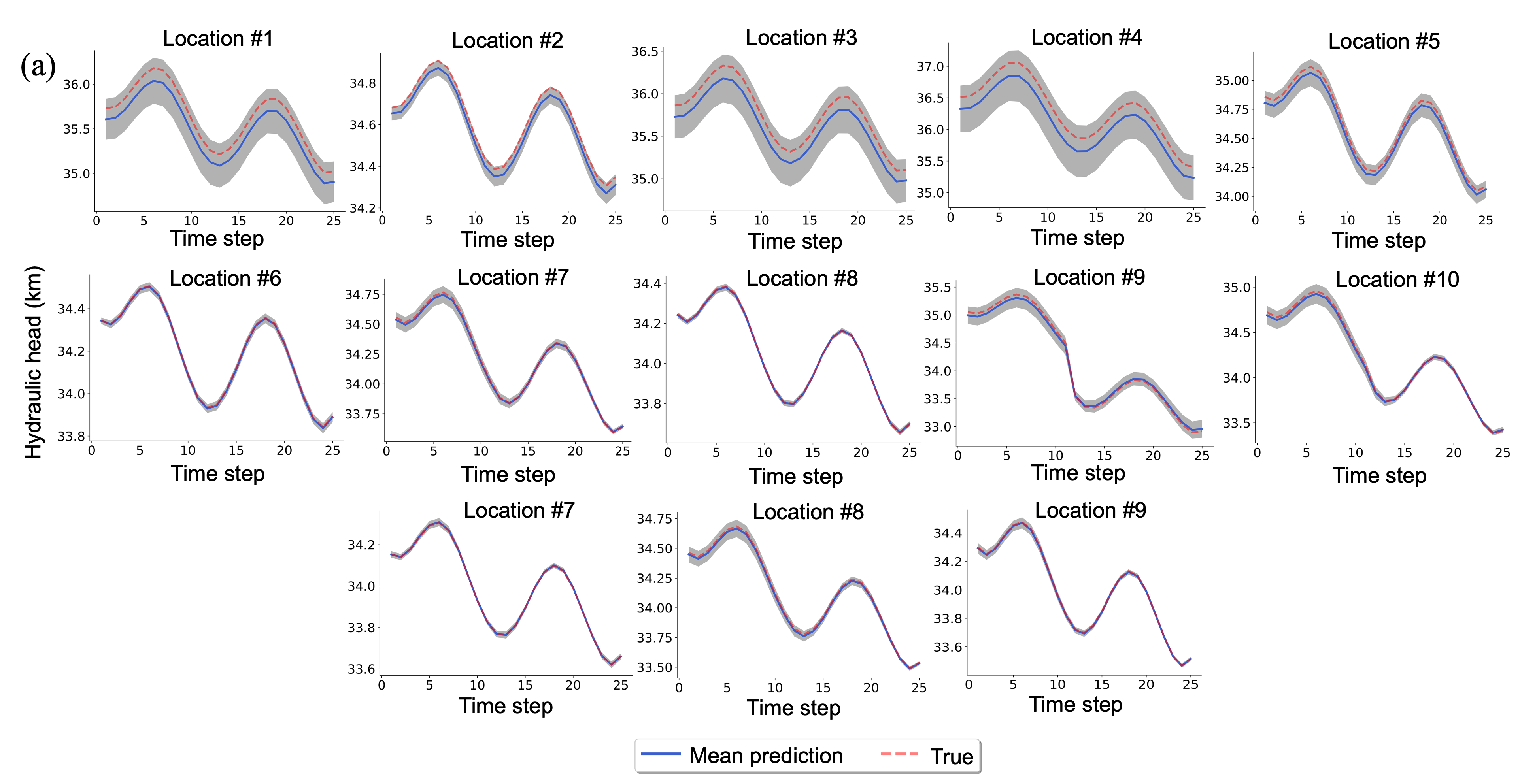}
  \end{subfigure}
  \begin{subfigure}[b]{\textwidth}
    \includegraphics[width=\textwidth]{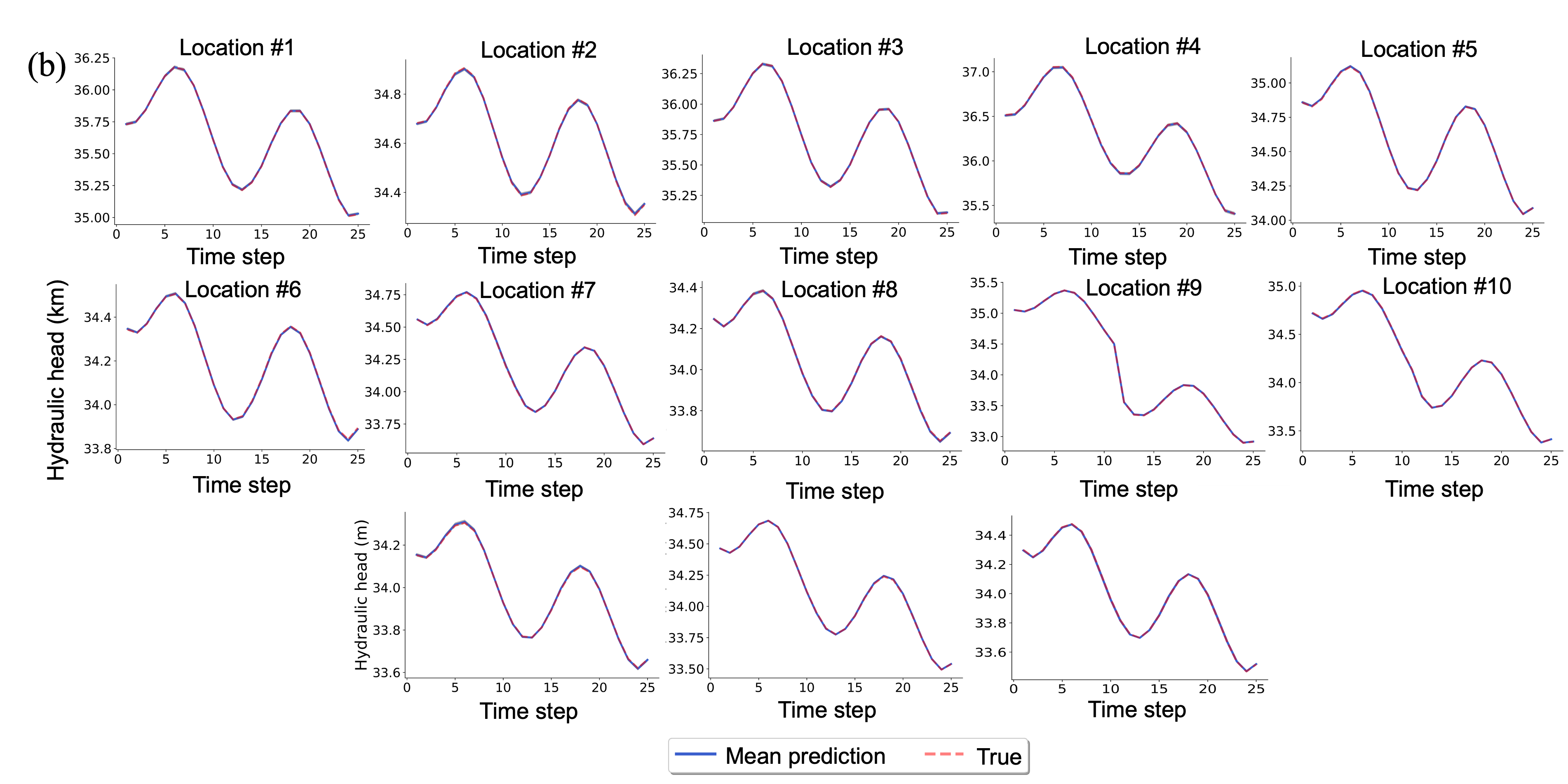}
  \end{subfigure}
  
  \caption{Observation prediction at 13 locations for test \#2: (a) proposed method; (b) PEST-IES.}
  \label{fig:17}
\end{figure}

In this section, we evaluate the posterior predictive distribution of hydraulic head measurements obtained using PEST-IES and our proposed method. The posterior predictive distribution is calculated by inputting the estimated log-conductivity field into the MODFLOW model. The proposed method is used to generate 2,000 posterior samples for each parameter, which are fed into the MODFLOW model, resulting in the same number of observation predictions.
Figs. \ref{fig:16} and \ref{fig:17} display the predictive observation distribution obtained via PEST-IES and the proposed methods, with the grey shaded areas representing the corresponding \(95\%\) credibility intervals. It can be seen that both methods are capable of accurately predicting observations across all locations and timesteps. For the PEST-IES method, the predictions closely coincide with the true values for both datasets with a notably small uncertainty, as indicated by the tight credibility intervals. In contrast, the proposed method exhibits larger uncertainties, and the credibility intervals span a wider range than those of PEST-IES, especially for observations at locations \#1 through \#4. This difference can be attributed to the fact that PEST-IES performs parameter estimation using a forward model that directly generates the synthetic observation data, thus bypassing the need for model training.
Consequently, PEST-IES can yield more accurate predictions, albeit with the trade-off of requiring more computational resources due to the repeated evaluation of the forward model.

In the proposed method, the errors due to the summary and inference network training contribute to the parameter estimation error.
There are several error sources in the network models. Firstly, the optimization of the loss function inevitably introduces some error due to the finite number of training data, as described in Eq. \ref{eq:objective_empirical}. This error arises from the stochastic nature of the sampling process used during training.
Secondly, the summary network choice might be suboptimal, potentially failing to capture all crucial information. Additionally, the designed invertible network might not sufficiently capture the complexities of the forward model's behavior. For instance, the depth of the inference network in our study, which comprises 10 invertible blocks, may be inadequate for handling the 706-dimensional parameter space. Thirdly, fine-tuning hyperparameters such as the dropout rate, learning rate, and optimization method presents a complex challenge, with limited guidance available for the ideal design and training of neural networks. These errors can propagate into the predictive performance, affecting the uncertainty in observation predictions. Despite these larger uncertainties, the mean predictions from the proposed method consistently align well with the true observations and remain within the confidence intervals, underscoring the method’s reliability. Moreover, for test \#2, both the uncertainty and deviation from the true observations are more pronounced than for test \#1 because the proposed method finds it more challenging to estimate parameters for test \#2---this is evidenced by the higher relative \(\ell_2\) error reported in Table \ref{tab1}.

The proposed method requires approximately 4.5 hours for training, but once this phase is complete, it can perform parameter estimation for a given measurement dataset in about 2 seconds. In contrast, PEST-IES requires around 2.5 hours to complete parameter estimation for each measurement dataset, as the IES algorithm must be restarted from scratch for each measurement dataset. Although the proposed method demands more time for training, this process can be conducted offline using simulated data. 
The amortized nature of the proposed approach allows for real-time inference and is valuable for parameter estimation problems with varying parameters and excitations that have to be estimated from new measurement data, such as varying boundary conditions to be estimated from time-series pressure observations.

\section{Conclusion}

In this work, we introduce a NF-based likelihood-free framework for tackling high-dimensional inverse problems. The proposed method includes three phases. The first phase involves generating synthetic training data from a forward model utilizing prior knowledge of the model parameters. In the second phase we train jointly the model's inference and summary networks. The summary network autonomously learns the most informative representation of the observation data, ensuring that the features extracted are maximally informative for inference tasks. The inference network is designed using an architecture that combines cACL and SL layers. The summary features are passed to the inference network, allowing for the joint optimization of the two networks. The final phase consists of real-time inference. Once the model is trained, the posteriors of the model parameters can be computed in real time. We apply this framework to estimate the 706-dimensional hydraulic conductivity field using noisy and incomplete observations from the Freyberg groundwater model. Additionally, we conduct a comparative analysis between our method and the likelihood-based benchmark method, PEST-IES. The results highlight that our method can accurately and efficiently estimate the high-dimensional conductivity field. Below, we summarize the main advantages of our proposed method.

First, the proposed method represents a fully likelihood-free approach that directly estimates the posterior without the need to evaluate a likelihood function and does not rely on assumptions on the structure of the true posterior beyond the expressivity of the inference network.
Second, the method exhibits robust performance across varied measurement sequences. In other words, it can handle variable dataset sizes with a single trained model, that is, without the need of re-training for different measurement sequences. This flexibility is particularly advantageous in practical settings when varying parameters must be estimated from new measurement data.
Third, the method includes a learnable summary network that effectively reduces data dimensionality from potentially large time-series datasets to a fixed-size vector. Unlike traditional likelihood-based methods, where summary statistics are manually selected by practitioners, our method autonomously learns the most informative features from raw data. 
Lastly, the proposed method offers significant computational advantages, particularly for problems that necessitate multiple runs the forward model from scratch given new datasets. This efficiency stems from the method's implementation of amortized inference, which provides a clear benefit over ABC and traditional sampling-based methods. While the training phase has a relatively high computational cost, this phase can be conducted offline. Once trained, the model can efficiently estimate the posterior distribution.

\bibliographystyle{unsrt}
\bibliography{reference}

\end{document}